\documentclass{article}

\usepackage[final]{corl_2026} %

\usepackage[numbers]{natbib}
\usepackage{multicol, multirow}
\usepackage{booktabs}
\usepackage{amssymb}
\usepackage{pifont} %
\usepackage{graphicx}
\usepackage{subcaption}
\usepackage{bbm}
\usepackage{float}
\usepackage[font=small]{caption}
\usepackage{pifont}
\usepackage[dvipsnames]{xcolor}
\usepackage{xcolor}
\usepackage{fontawesome}
\usepackage{amsmath, algorithm, algpseudocode}
\usepackage{adjustbox}
\usepackage{wrapfig} %
\usepackage{colortbl} %
\usepackage{array}    %
\newcommand{\cmark}{\ding{51}}%
\newcommand{\xmark}{\ding{55}}%

\newcommand{\newaddition}{\textcolor{black}{}\textcolor{black}}

\title{SILO: Simulation-in-the-Loop Sim-to-Real Transfer for Multi-Stage Cable Routing}

\author{Stone Tao$^{1,2}$, Jie Xu$^1$, Hesam Rabeti$^1$, Yashraj Narang$^1$, Yijie Guo$^{1*}$, Iretiayo Akinola$^{1*}$\\$^1$NVIDIA Corporation, $^2$University of California, San Diego}

\begin{document}
\maketitle

\begin{abstract}
    Linear-deformable manipulation remains challenging due to the complex deformations of objects such as cables and ropes. Prior data-driven approaches, particularly imitation learning, have shown some promise in narrowly defined settings but typically require thousands of demonstrations for specific tasks and cable types, limiting scalability and generalization. We introduce a sim-to-real reinforcement learning (RL) framework for multi-stage cable routing that leverages GPU-parallelized simulation to approximate linear deformable behaviors. Training across thousands of parallel simulations enables the learned policies to generalize across diverse cable geometries and deformation patterns. To bridge the sim-to-real gap, we propose a novel deployment strategy that combines a \textbf{S}imulation \textbf{I}n the \textbf{LO}op (SILO) execution framework, localized RL policies, and robust cable state estimation. On real-world cable routing tasks, our approach achieves higher success rates and $\sim$2$\times$ reduction in cycle times compared to prior state-of-the-art learning methods. To our knowledge, this is the first successful sim-to-real transfer of RL policies for multi-stage cable routing. Videos and additional visualizations are available at \href{https://silo-cable-routing.github.io/}{https://silo-cable-routing.github.io/}
\end{abstract}

\keywords{Sim-to-Real, Cable Routing, Reinforcement Learning} 

\section{Introduction}

Industrial automation remains a central objective in robotics, yet its success has largely been confined to rigid body assembly where dynamics are predictable and execution can be governed by scripted trajectories or perception-driven motion planning pipelines. However, many real-world assembly processes, such as automotive wiring, hose routing, and textile handling require manipulating deformable objects. The high variability and dimensionality of deformable materials demand continuous, dexterous low-level control, posing a challenge for traditional robotic frameworks.

In this paper, we address the cable manipulation problem by learning reactive low-level policies with reinforcement learning (RL) in simulation followed by sim-to-real deployment. We focus on \textbf{cable routing}, a ubiquitous industrial task that requires handling subtle, local deformations during execution. These deformations are difficult to predict from visual observations alone and can change abruptly due to contact, gravity, and accumulated slack. As a result, cable routing requires policies that can react to sudden changes in cable deformations. Traditional systems often require substantial task-specific engineering to operate reliably in tightly constrained settings. Moreover, the unpredictable nature of deformable objects necessitates frequent replanning, making real-time, high-performance execution difficult to achieve.

\begin{figure}[t]
    \centering
    \includegraphics[width=\textwidth, clip]{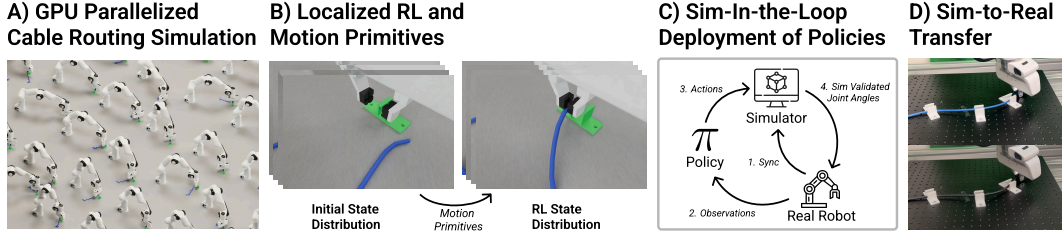}
    \caption{
    \textbf{Key components of our system.} \textbf{A)} GPU-parallelized articulated rigid-body simulation approximates cable dynamics during routing. \textbf{B)} Motion primitives grasp and transport the simulated cable to randomized intermediate states near the harness to efficiently train localized RL policies. \textbf{C)} At deployment, the robot executes identical motion primitives before transitioning to simulation-in-the-loop execution. Here, the RL policy actions are executed in simulation, and the real robot tracks the resulting simulated joint angles, reducing controller dynamics gap and avoiding harness collisions. \textbf{D)} Our system enables sim-to-real transfer for multi-stage cable routing, achieving higher success rates with $\sim2\times$ lower cycle times compared to baselines.}
  \label{fig:system_overview}
  \vspace{-2.0em}
\end{figure}

As an alternative to traditional planning based systems, data-driven methods present a more general approach to train robotics policies. Imitation learning based approaches \citep{chi2023diffusionpolicy, DBLP:conf/rss/ZhaoKLF23} leverage task-specific data, and in some cases large pre-training datasets are used in VLAs \citep{black2410pi0, pmlr-v270-kim25c, gr00tn1_2025}. While imitation learning is fairly general, it requires sufficient high-quality demonstrations that can be costly or impractical to collect for deformable object tasks. The work most related to ours is the hierarchical imitation learning approach \citep{luo2023multistage} that demonstrated promising results for multi-stage cable routing, however it requires 1000s of demonstrations. In contrast, RL-based sim-to-real approaches require no demonstrations and train policies via scalable simulation followed by deployment. However, successful sim-to-real RL has largely been limited to rigid body manipulation \citep{noseworthy2025, tang2023} or locomotion \citep{tairanhe2025, rudin2021}. Effective RL requires both high-throughput simulation for fast training as well as an accurate enough simulator for sim-to-real transfer. While these requirements can be satisfied for rigid-body systems, traditional soft-body simulators for objects such as cloth or cables typically trade off speed for physical fidelity.

A key insight of our work is that linear deformables can be approximately simulated using GPU-parallelized rigid-body physics. Furthermore, we introduce SILO (\textbf{S}imulation \textbf{I}n the \textbf{LO}op), a deployment framework that continuously synchronizes the robots in simulation and in real to mitigate controller dynamics gaps. Combined with a state estimation system for linear deformables, we can learn reactive low-level policies that enable faster and more reliable cable routing than prior state of the art \citep{luo2023multistage}. An overview of our system is shown in Fig 1. Our primary contributions are as follows: 

\textbf{GPU-parallelized Simulation for Linear-Deformables} that approximates cable deformation using rigid-body articulations, enabling large-scale RL training. 

\textbf{Localized RL for Cable Routing} that decomposes the task to focus RL on routing behaviors that address complex deformations. Motion primitives handle grasping and free-space motion.

\textbf{Robust State Estimation} that fits polylines onto segmented cable point clouds, producing linear deformable observations matching the simulated cable discretization for sim-to-real transfer.

\textbf{Simulation-In-the-Loop (SILO)} deployment paradigm that reuses simulation dynamics at runtime, enabling RL policies trained in simulation to be executed on real robots without system ID.

Together, these contributions form a novel system that achieves higher success rates in multi-stage cable routing and a $\sim$2$\times$ reduction in cycle times compared to prior work \citep{luo2023multistage}. Our approach generalizes across 4 cable types varying in materials and stiffness, as well as initial harness configurations without requiring demonstrations.

\section{Related Work}
\textbf{Deformable Manipulation:}
Past work tackling deformable manipulation can be split into three categories: motion planning, imitation learning, and reinforcement learning. 

Planning-based approaches combine perception stacks with motion primitives to tackle cable routing. However, these methods often require a lot of manual tuning and engineering in both designing hardware and plans \citep{DBLP:conf/case/AzulayKDXLCG25, DBLP:conf/iros/0005BWMKHK23, DBLP:journals/ral/JinLWTS22, DBLP:conf/icra/WalterssonLK22, chen2026craft, DBLP:conf/rss/DongWSSRA20, DBLP:journals/corr/abs-2510-19268}. Moreover they often assume full observability \citep{DBLP:conf/icra/SchulmanLHA13, DBLP:journals/ral/YanZJB20}. \newaddition{Furthermore, trajectory optimization approaches for motion generation often use highly-simplified dynamics models for cable modeling \citep{DBLP:conf/icra/JinRRJT21}.} These constraints cause planning approaches to overfit and have difficulty generalizing to new settings. Furthermore, the planning nature further limits the speed of these systems.

Imitation learning (IL) methods \citep{chi2023diffusionpolicy, DBLP:conf/iros/0002OK24, DBLP:conf/rss/ZhaoKLF23} have shown some evidence of tackling deformable manipulations like cable routing \citep{DBLP:conf/rss/ZhaoKLF23}. \newaddition{Compared to planning-based approaches, learning-based approaches demonstrate a higher level of generalizability while requiring less engineering and manual tuning}. However, they rely on massive datasets of demonstrations and require task-specific finetuning. A hierarchical IL approach has been proposed to tackle the cable routing problem \cite{luo2023multistage}, which is the most related work to this paper. However the hierarchical approach relies on 1000s of demonstrations to achieve some success, while our work does not need any demonstrations and relies on RL.

Most past work on RL for deformable manipulation stay within simulation without evidence of sim-to-real transfer \citep{DBLP:conf/iclr/HoangLBVN25, DBLP:conf/corl/0002ZWGSMWLLSAH22, DBLP:conf/corl/LinWOH20, DBLP:conf/iclr/XingLO25}. Some research has explored both state-based \citep{DBLP:journals/ral/WengZYKVNK24} and vision-based sim-to-real deformable manipulation \citep{DBLP:conf/corl/MatasJD18, DBLP:journals/ral/ScheiklTGWDFM23, DBLP:conf/rss/WuYKPA20}. However, past sim-to-real work on deformables has been limited significantly in terms of generalizability and complexity, in part due to slow simulation and/or inaccurate dynamics. They often can only tackle simpler tasks where there are only simple interactions between just the gripper and the deformable object \citep{DBLP:journals/ral/ScheiklTGWDFM23, DBLP:conf/rss/WuYKPA20}. Our work tackles a cable routing problem that is dynamically complex via GPU simulation and sim-to-real RL.

\textbf{Simulation:}
There are a variety of simulation frameworks that provide support for both rigid body physics and/or soft body physics, such as PyBullet \citep{coumans2019}, PhysX \cite{NVIDIA_Isaac_Sim}, Mujoco \citep{todorov2012mujoco}, and Newton \citep{newton}. Prior works have leveraged these physics engines to simulate diverse tasks for robotics. SAPIEN/ManiSkill \citep{DBLP:conf/cvpr/XiangQMXZLLJYWY20, gu2023maniskill2, taomaniskill3} is built on top of PhysX and Warp, providing a suite of simulated robotic learning benchmarks. Rewarped \citep{DBLP:conf/iclr/XingLO25} uses Warp \citep{warp2022} for one-way coupled soft body functionalities; Plasticinelab \citep{DBLP:conf/iclr/HuangHDZ0TG21} adopts Taichi \citep{hu2019taichi} to simulate manipulations of deformables. Previous works have explored speeding up the dynamics simulation of deformables using rigid body approximation. \citet{DBLP:conf/icra/LiASB23} leverage a CPU-based rigid-body simulator \citep{xu2021diffhand} to approximate the dynamics of a rope. Similar to our work \citet{DBLP:journals/ral/JacobBWBR24} reuse existing parallelized rigid body APIs via Isaac Gym \citep{makoviychuk2021isaac} for tree branch simulation. In our paper, we leverage the PhysX GPU parallelized rigid body APIs via ManiSkill3 to approximate linear-deformable dynamics.

\textbf{Sim-to-Real Transfer:} The sim-to-real gap in both dynamics and visuals makes sim-to-real difficult. To address sim-to-real gaps, past work has leveraged state estimation \citep{jiang2024transic} or pointcloud data \citep{christen:cvpr2023, DBLP:conf/corl/QinHY0022}. Other work has leveraged style adaptation in image space to make simulation look more realistic \citep{DBLP:journals/ral/ScheiklTGWDFM23}. Image domain randomization trains policies to be invariant to simulation image artifacts and transfer to the real world better \citep{DBLP:conf/corl/MatasJD18, DBLP:conf/iros/TobinFRSZA17}. With GPU parallelized simulators, past work has shown how large-scale domain randomized images \citep{DBLP:journals/corr/abs-2412-01791, taomaniskill3, mujoco_playground_2025} and physical properties \citep{mittal2025isaaclab, rudin2021}, enable faster and more generalizable sim-to-real transfer of RL policies for both locomotion and manipulation of rigid bodies. Our work trains RL policies via a GPU simulation designed to approximate linear deformable dynamics. To the best of our knowledge, our work is the first to demonstrate zero-shot sim-to-real transfer of an RL policy for deformable cable routing. 

\section{Problem Description}
\newaddition{The goal is to solve the long-horizon task of routing a linear deformable cable through a sequence of $N$ harnesses. We use RL-based sim-to-real transfer for task components requiring reactive adaptation to deformations (i.e., routing), while predefined motion primitives handle grasping and free-space motion. We assume harness geometries and poses are known a priori, either from CAD models or a perception pipeline.} Let $h_i \in SE(3)$ be the pose of the $i$th harness. For a given type of harness (a U-shaped clip in our work), we define a fixed transformation $T$ that maps the harness pose to $p_i = Th_i$, where $p_i$ is an end-effector pose positioned behind the harness prior to routing. The orientation of $p_i$ disambiguates the routing direction for harnesses with geometric symmetries. The ordered set $\{(h_i, p_i)\}_{i=1}^N$ defines a cable routing instance in our formulation.

\section{Methodology}
\subsection{Simulation Design}
\begin{wrapfigure}{r}{0.4\textwidth}
    \vspace{-1.0em}
    \includegraphics[width=1\linewidth]{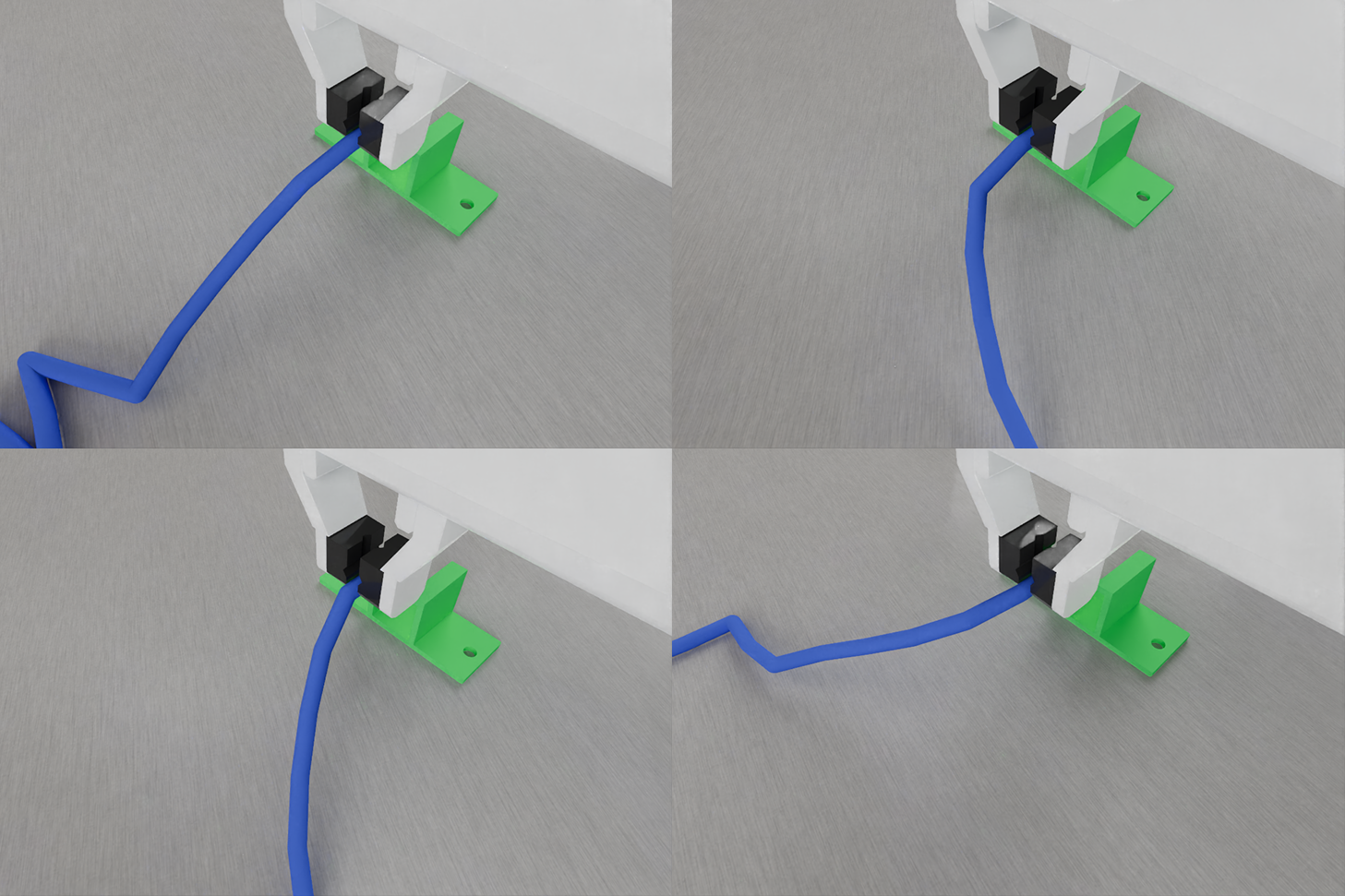}
    \caption{Four example intermediate states of our cable simulation generated by motion primitives for initializing episodes during RL training. The simulated Franka robot gripper is grasping the end of a cable and is placed just behind a harness at pose $p_i$.}
    \label{fig:intermediate_state_reset_distribution}
    \vspace{-1.5em}
\end{wrapfigure}
Our key design choice is to model linear-deformables using only rigid-body APIs, trading off fidelity for high-throughput simulation to enable fast RL training. We leverage the rigid-body APIs in ManiSkill3 to approximate linear-deformables that exhibit elastic and plastic behaviors. Concretely, we model a linear-deformable as an articulated sequence of links. Each link is modeled visually and physically as a rigid capsule. Adjacent links are connected by 3 orthogonal revolute joints, allowing the cable to bend freely in 3D.

\newaddition{To simulate cable behaviors, we apply two operations after each physics timestep. Elasticity is modeled by driving each joint towards a saved rest position using standard PD control parameterized by joint stiffness and damping. Plasticity is modeled by gradually updating each joint’s rest position toward its current configuration. By tuning the PD gains and plasticity coefficients, this formulation captures a range of behaviors spanning highly elastic cables (e.g., coaxial cables) to more plastic, rope-like materials. While this model does not correspond to typical material parameters, it provides a computationally inexpensive approximation that achieves sufficient state coverage for the cable-routing task and supports successful sim-to-real transfer. Visuals of the environment are shown in Fig. \ref{fig:intermediate_state_reset_distribution}, with more details on the simulation design and parameters in Appendix \ref{appendix:sim-design}.}

\subsection{Reinforcement Learning Setup}

\label{sec:rl_training_intermediate_states}

\textbf{Problem Localization with Intermediate State Resets:} We train our cable manipulation policy using Proximal Policy Optimization \cite{schulman-ppo}. The full task consists of grasping the cable tip and sequentially routing it through multiple harnesses. To reduce task complexity and improve training efficiency, we localize RL to a single subtask: routing a grasped cable through one harness. Localization shortens episode horizons and simplifies credit assignment, in addition to simplifying reward design. Furthermore, simulation cost is reduced by allowing the use of shorter cables since we only route through one harness with RL. To achieve this localization, before RL training, we generate a dataset of $n$ intermediate simulation states $\mathcal{D}=\{e_1, .., e_n\}$ where the robot has already grasped the cable and is positioned at a pose $p_i$ just behind the harness to route through. A set of motion primitives are used to achieve these states in simulation (see Section \ref{sec:motion_primitives}).
During RL training, environment resets sample uniformly from $\mathcal{D}$, visualized in Fig. \ref{fig:intermediate_state_reset_distribution}.

\textbf{Reward Function:}
Success is defined by the proximity of the two cable links closest to the gripper, $l_1$ and $l_2$, to the harness center $h_{\text{pos}}$ within a radius $\epsilon$. The reward function at time $t$ is defined as: \\$r_t = \exp(-c||l_1 - h_\text{pos}||_2) + \exp(-c||l_2 - h_\text{pos}||_2) +\mathbbm{1}_\text{grasped} + \mathbbm{1}_{\text{success}}$
where $c$ is a scaling coefficient. The indicator $\mathbbm{1}_\text{grasped}$ equals $1$ when the cable is securely grasped. $\mathbbm{1}_{\text{success}}$ is $1$ when success is achieved. While more elaborate reward shaping is possible, we find this simple formulation is sufficient and leads to emergent routing behaviors (visualized in Appendix \ref{appendix:learned_behaviors_visualizations}). Hyperparameters and training curves are provided in Appendix \ref{appendix:reinforcement_learning}.

\subsection{Sim-to-Real Transfer}

\begin{figure}
    \includegraphics[width=\linewidth]{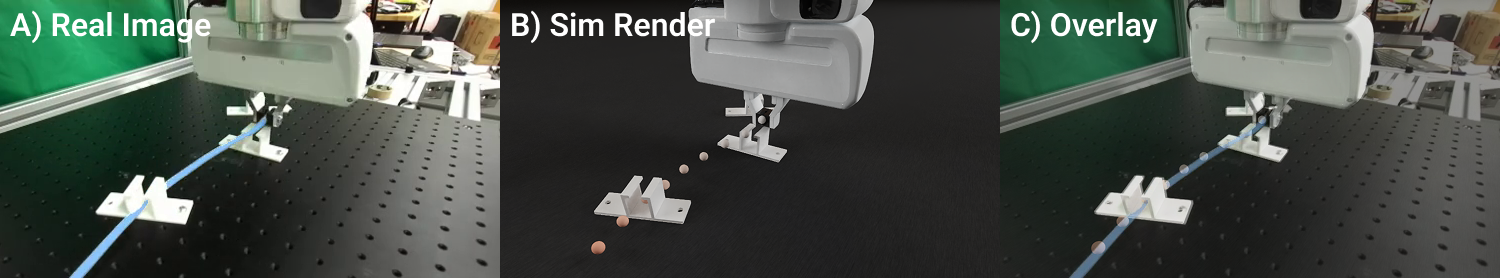}
    \caption{Three images showing A) the real world, B) rendering of the digital twin, C) an overlay of the task from the Zed 2 camera's perspective. The blue shaded area in A) and C) denotes the segmentation of the cable generated by SAM2. The spheres in the B) and C) represent the post-processed estimated cable point sequence.}
    \label{fig:sim-to-real_obs}
    \vspace{-2.2em}
\end{figure}

\begin{wrapfigure}[9]{r}{0.42\textwidth}
    \vspace{-1em}
    \includegraphics[width=\linewidth]{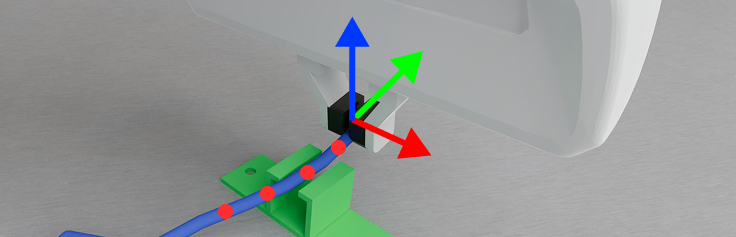}
    \caption{Observation features in simulation. 
    The red dots denote the 4 observed cable points closest to the TCP. The RGB axes show the observed TCP pose.}
    \label{fig:obs_space_design}
\end{wrapfigure}

\textbf{Observations:}
The sim-to-real problem for linear deformables is relatively under-explored, making the dynamics gap a central challenge. This motivates us to adopt state-based rather than vision-based policies to isolate the dynamics gap. Thus, we develop a robust state estimation pipeline that enables state-based policies in both simulation and the real world. Concretely, the observation consists of the positions of four cable points immediately behind the gripper, the tool-center-point (TCP) pose, and robot joint angles, visualized in Fig. \ref{fig:obs_space_design}. Gaussian noise is added to the cable points during training.

\newaddition{In simulation, cable points are obtained directly from the link states, where each link has equal length. In the real world, in order to estimate sim-to-real compatible cable points, we fit an equal-length polyline to a segmented cable pointcloud. To predict the segmented cable pointcloud, a depth image is generated using Foundation Stereo \citep{wen2025stereo} from stereo RGB images produced by a Zed 2 scene camera. The Segment Anything Model 2 \citep{ravi2025sam} is used to predict a cable segmentation mask, and the segmented depth is projected to the world frame using calibrated camera parameters. Fig. \ref{fig:sim-to-real_obs} shows example state-estimation results, with additional details in Appendix \ref{appendix:state_estimation}.}

\textbf{Motion Primitives:}
\label{sec:motion_primitives}
The full task is decomposed into structured segments handled by motion primitives and the RL policy. We define two motion-planning-based primitives, \textit{GraspCable} and \textit{MoveToHarness}, which are used both for generating the intermediate states and for physical deployment.

\textit{GraspCable}: The robot approaches a cable endpoint from above and closes the gripper. In simulation, the cable endpoint is known exactly; in the real world, it is obtained via state estimation. Motion planning then generates a sequence of joint positions for this primitive which is executed once per trial; once the cable is secured, the gripper remains closed until task completion or failure.

\textit{MoveToHarness}: The motion planner is used to move the robot's TCP pose to the computed $p_i$ behind harness $i$, oriented for routing.

In simulation, episodes are localized to a single harness. To generate a diverse training distribution, we initialize each episode by sampling a harness-pose tuple $(h_i, p_i)$. The system then executes \textit{GraspCable} followed by \textit{MoveToHarness} to position the TCP at $p_i$. This automated sequence generates the intermediate state distribution illustrated in Fig. \ref{fig:intermediate_state_reset_distribution}, allowing the RL agent to begin its learning process directly at the start of the difficult routing maneuver. During deployment, the user provides a sequence of harness configurations $\{(h_1, p_1), ..., (h_n, p_n)\}$. The system first executes \textit{GraspCable}, then iterates through the sequence. For each iteration $i$, the robot executes \textit{MoveToHarness} to reach $p_i$, at which point control is handed over to the RL policy to perform the routing through $h_i$. Visualizations of the primitives are detailed in Appendix \ref{appendix:motion_primitives}.

\begin{wrapfigure}{r}{0.52\textwidth}
\vspace{-2.2em}
\begin{minipage}{0.52\textwidth}
\begin{algorithm}[H]
\caption{Simulation-In-the-Loop Deployment}
\label{alg:silo}
\begin{algorithmic}[1]
\Require Policy $\pi$, simulator $\textsc{Sim}$, real robot $\mathcal{R}$
\State Initialize simulator $\textsc{Sim}$ with real harness poses
\For{$t = 1$ to $T$}
    \State Read real robot joint positions $q^{real}_t$
    \State Set simulator joints $q^{sim}_t \leftarrow q^{real}_t$
    \State Estimate cable points $\hat{c}_t$ from perception
    \State Construct observation $o_t$ using $q_t^{sim}$ and $\hat{c}_t$
    \State Compute action $a_t \leftarrow \pi(o_t)$
    \State Step simulator with $a_t$
    \State Command real robot joints to move to $q^{sim}_{t+1}$
\EndFor
\end{algorithmic}
\end{algorithm}
\end{minipage}
\vspace{-1.0em}
\end{wrapfigure}

\textbf{Sim-to-Real Deployment:} To mitigate the sim-to-real dynamics gap, we propose \textbf{SILO:} \textbf{S}imulation \textbf{I}n the \textbf{LO}op, a deployment framework that leverages the simulation’s internal dynamics as a reference for real-world execution. Traditional sim-to-real workflows often require extensive system ID to align simulation and real controllers. SILO bypasses system ID by treating the simulator as the primary source of truth for robot motion during deployment. Rather than transferring policy actions directly to the real robot, we transfer simulated outcomes. During policy execution in deployment, we initialize a digital twin that mirrors the physical setup visualized in Fig. \ref{fig:sim-to-real_obs} by including harness geometries/poses and the robot. Unlike the training environment, we do not explicitly simulate a physical cable during deployment; instead, cable observations are generated via real-world perception.

SILO is described in Algorithm \ref{alg:silo}. At each RL execution timestep, we sync the simulated robot with the real robot's joint positions, generate observations, and query the policy for an action. Instead of interpreting the action with real-world controllers, the action is applied in simulation using the same controller and parameters used during training. The resulting joint angles from the simulation step are then used as joint targets for the real robot. The real controller moves the arm to these joint targets to re-sync the real and simulated states. This iterative process is outlined in Algorithm \ref{alg:silo} and illustrated in Fig. 1. We detail a few advantages SILO offers for sim-to-real deployment:

\textbf{Controller Agnosticism and Flexibility:} By using the simulator to generate joint targets, SILO eliminates the need for precise alignment between the simulated and real-world controller prior to sim-to-real transfer. In our experiments, we successfully deploy policies using an off-the-shelf real-robot controller without any tuning. \newaddition{Furthermore, SILO allows for flexibility in controllers by permitting the use of different controller parameters (e.g., PD gains) that can sufficiently achieve joint targets. For controller implementation details, see Appendix \ref{appendix:controller}.}

\textbf{Inherent Safety and Collision Avoidance:}
Since actions are first executed in a digital twin of the real environment, physical constraints such as intersection-free states are enforced before commands are sent to the real robot. As a result, the real robot receives collision-free joint targets, reducing the reliance on handcrafted RL reward shaping terms such as collision penalties.

\textbf{Reduced Engineering Overhead:}
SILO allows the same simulation codebase to be reused for training, observation generation, and deployment. This reduces duplicated logic, minimizes implementation errors, and simplifies debugging and visualization during real-world execution.

Implementation details and further visualizations of SILO are provided in Appendix \ref{appendix:silo}.

\section{Experiments}
\begin{wrapfigure}{r}{0.425\textwidth}
    \vspace{-5.0em}
    \centering
    \includegraphics[width=1\linewidth]{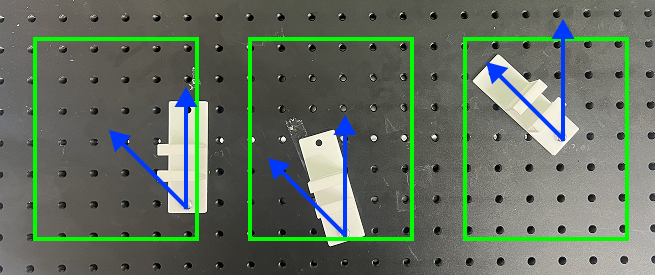}
    \caption{
    Sampling distribution of harness poses for cable routing. 1 to 3 harnesses are independently sampled within one of three {\color{ForestGreen} $12.5\text{cm} \times 15\text{cm}$ regions}, with {\color{blue}in-plane rotations uniformly sampled up to $45^\circ$}. The green boxes are spaced out by 5 cm.}
    \label{fig:baseline_benchmark_reset_dist}
    \vspace{-1.5em}
\end{wrapfigure}
We evaluate on a cable routing benchmark consisting of routing a single cable through one to three harnesses. We follow the experimental protocol of \citet{luo2023multistage}, which uses a fixed harness geometry and cable type (nylon rope), while varying harness poses sampled independently from a predefined spatial distribution (Fig. ~\ref{fig:baseline_benchmark_reset_dist}). Through real-world experiments, we aim to answer the following research questions:

\textbf{Zero-shot sim-to-real performance:} Can RL policies trained in simulation achieve reliable zero-shot deployment for multi-stage cable routing? How does it compare to prior data-driven methods?

\textbf{Generalization:} How robust is the RL policy to variation in cable and harness configurations?

\textbf{SILO Deployment}: How does SILO perform compared to other deployment methods in terms of performance and flexibility?

\subsection{Main Performance Results}

\textbf{Performance}: Table~\ref{table:main_results} summarizes success rates and cycle times for routing up to 3 harnesses compared to scripted and hierarchical imitation learning (h-IL) baselines, both reported in prior work \citep{luo2023multistage}. SILO achieves higher success across all settings. Failures with multiple harnesses primarily occur in configurations with large relative orientation changes between harnesses that induce extreme cable entry angles. We analyze this effect in detail through controlled ablation studies in Section \ref{sec:ablations}. Furthermore, our system achieves much lower cycle times. While h-IL relies on planning to reshape cables in high-curvature states, our RL policy handles them directly, eliminating frequent regrasping and planning iterations in h-IL, leading to $\sim$$2\times$ faster cycle-times. See Appendix \ref{appendix:cycle_times} for a cycle-time breakdown.

\begin{wraptable}{r}{0.47\textwidth}
    \small
    \vspace{-2.7em}
    \begin{tabular}{ccccc}
        \toprule
         & H1 & H2 & H3 & Time (s) \\
         \midrule
         Scripted  & -- & -- & 5/24 & -- \\
         h-IL & 19/24 & 14/24 & 12/24 & $\sim$200\\
         SILO & \textbf{24/24} & \textbf{22/24} & \textbf{18/24} & 87.48\\
         \bottomrule
    \end{tabular}
    \caption{Success rates and execution times for cable routing. H1-H3 denote routing through 1 to 3 harnesses. Reported solve time corresponds to routing through all three harnesses. Baseline results (Scripted and h-IL) are copied verbatim from prior work. Time for the h-IL baseline is estimated from published videos. ``--'' indicates unavailable data.}
    \label{table:main_results}
    \vspace{-1.5em}
\end{wraptable}

While baseline results are reported verbatim from \citet{luo2023multistage}, the comparison is not strictly apples-to-apples. The h-IL system assumes a fixed cable type and relies on explicit replanning and reshaping primitives, while our approach assumes known harness poses and geometries. Thus, we view the methods as addressing different sides of the problem: h-IL emphasizes planning and recovery, whereas our approach demonstrates reactive low-level routing via sim-to-real RL. Combining the high-level planning in h-IL with our RL routing is a promising direction we leave to future work.

\begin{wraptable}{r}{0.63\textwidth}
    \small
    \vspace{-1.2em}
    \centering
    \begin{tabular}{cccc}
        \toprule
         Nylon Rope & Ethernet Cable & Charger Cable & HDMI Cable \\
         \midrule
         18/24 & 18/24 & 18/24 & 14/24 \\
         \bottomrule
    \end{tabular}
    \caption{Success rates for routing different cables through 3 harnesses.}
    \label{table:different_cable_results}
    \vspace{-1.0em}
\end{wraptable}

\textbf{Generalization}: Table~\ref{table:different_cable_results} reports success rates when routing with \textit{different} real-world cables through three harnesses. Performance remains consistent across different cable types, indicating robustness to substantial variation in stiffness and material. The thicker HDMI cable introduces additional failures, suggesting that cable diameter becomes a limiting factor when the clearance between harnesses is small. Overall, these results demonstrate strong zero-shot generalization across cable types. See Appendix \ref{appendix:cable_properties} for pictures and properties of the cables.

\textbf{Learned Strategies}: \newaddition{Analysis of real-world rollouts reveals two emergent strategies learned by RL in simulation that were not explicitly programmed in the reward function. We find that the policy exhibits an ``angling'' behavior where the gripper is oriented to drive the cable closer to the harness opening. Moreover, the policy exhibits a reactive ``swinging'' behavior. Depending on the cable's curvature, the policy will swing to one side of a harness to facilitate successful routing. See our website for videos of these strategies or still frames in Appendix \ref{appendix:learned_behaviors_visualizations}}.

\subsection{Ablations}

\begin{wrapfigure}{r}{0.39\textwidth}
    \vspace{-5.2em}
    \centering
    \includegraphics[width=\linewidth]{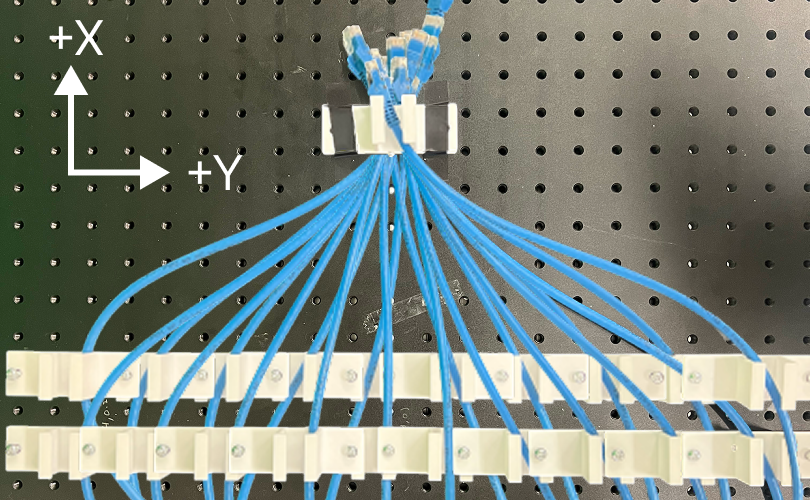}
    \caption{Reset states used to densely evaluate trained policies. The cable is shown already routed through the target harness (top) to show what a solved state looks like.}
    \label{fig:ablation_reset_dist}
    \vspace{-1.5em}
\end{wrapfigure}
\label{sec:ablations}

To better understand what enables our method to bridge the sim-to-real dynamics gap, we conduct ablation studies on our simulation environment. To keep tests controlled and fast, we analyze the performance of our policy in routing through a single harness. For each tested RL policy, we perform evaluations on a fixed set of 20 start states, visualized in Fig  \ref{fig:ablation_reset_dist}. Given that cables can easily deform, we control the starting states by first routing them through a harness (bottom of Fig \ref{fig:ablation_reset_dist}) which is affixed to 1 of 20 spots. The 20 start states cover cable entry angles ranging from $-60^\circ$ to $60^\circ$, relative to the target harness. Similar to \citep{luo2023multistage}, we observe that large entry angles are one of the main reasons for failures in cable routing, motivating our experimental setup here.

\begin{wraptable}{r}{0.5\textwidth}
    \small
    \vspace{-1.5em}
    \centering
    \begin{tabular}{ccccc}
        \toprule
         Randomization (m) & 0.0 & 0.10 & 0.20 & 0.40 \\
         \midrule
         Success & 5/20 & 5/20 & 8/20 & \textbf{15/20} \\
         \bottomrule
    \end{tabular}
    \caption{Success rate of policies trained with increasing randomization of cable states. Randomization numbers are the amount of translation in meters along the surface's y-axis that the cable can be randomized around.}
    \label{fig:intermediate_state_randomization_ablation}
    \vspace{-1em}
\end{wraptable}
\textbf{Intermediate State Sampling:}
We ablate the amount of initial state randomization used when sampling intermediate training states. Specifically, we vary the magnitude of translation along the board’s $y$-axis, which directly controls the cable’s entry angle into the target harness. As shown in Table~\ref{fig:intermediate_state_randomization_ablation}, increased randomization substantially improves real-world success rates. Notably, without explicit randomization, the policy succeeds in a few trials, likely due to stochasticity induced by exploration during training. The best-performing policy fails only at the most extreme entry angles, highlighting both the benefits and remaining limitations of sim-to-real RL for highly contorted configurations. See Appendix \ref{appendix:example_failures} for visuals of failure modes.

\begin{wraptable}{r}{5.3cm}
    \vspace{-1em}
    \small
    \centering
    \begin{tabular}{cccc}
        \toprule
         Method & Contact Penalty & Success \\
         \midrule
         SILO & \xmark & \textbf{15/20} \\
         Tuned & \xmark & 0/20 \\
         Tuned & \cmark & 10/20 \\
         \bottomrule
    \end{tabular}
    \caption{Success rate of policies with different deployment methods and contact penalty inclusion. SILO + contact penalties are omitted because the SILO setup enforces collision avoidance.}
    \label{table:deployment_ablations}
    \vspace{-1.5em}
\end{wraptable}

\textbf{Deployment:} \newaddition{We compare SILO against a baseline that tunes simulation PD gains to match the real robot controller. Since the baseline lacks implicit collision avoidance, we add a contact penalty reward option to discourage contacting the table and harness. Results in Table \ref{table:deployment_ablations} demonstrate that SILO outperforms the baseline approach in terms of success rate. Moreover, SILO by nature does not need contact penalties, simplifying reward design. A key contributor to SILO's success rate is its reduced sensitivity to controller discrepancies; without SILO, even after carefully tuning the simulated controller to match the response of the real-world controller, there is still a mean error of 0.5 degrees per joint. For further details on the tuning procedure for the simulated controller, as well as a visualization of the controller dynamics gap, see Appendix \ref{appendix:sim-controller-tuning}.}

\begin{wraptable}{r}{6.2cm}
    \vspace{-1em}
    \small
    \begin{tabular}{lc}
        \toprule
         Controller & Success \\
         \midrule
         Delta Joint Position (Medium PD) & 15/20 \\
         \midrule
         Delta Joint Position (Low PD) & 14/20 \\
         \midrule
         Delta Joint Position (High PD) & 14/20 \\
         \bottomrule
    \end{tabular}
    \caption{Success rate of policies trained with different simulation controllers.}
    \label{table:controller_ablations}
    \vspace{-1em}
\end{wraptable}
\textbf{Controller:} \newaddition{To demonstrate the flexibility of the SILO deployment approach, we train a policy on the same task, but with different simulation controllers. Specifically, in simulation, we vary the PD values for our delta joint position controller between three modes: low, medium, and high. The medium setting is the default used for all other experiments. Results shown in Table \ref{table:controller_ablations} show that SILO allows zero-shot deployment of a variety of controller settings, all without any additional manual tuning that would be required by non-SILO approaches. Details on the low-level simulation controller are provided in Appendix \ref{appendix:sim-controller}.}

\section{Conclusion and Limitations} 
\label{sec:conclusion}

Our work introduces a novel multi-stage cable routing system that leverages approximated GPU parallelized linear deformable dynamics, localized reinforcement learning, state estimation, and SILO deployment for more robust and efficient cable routing. Through experiments and ablation studies, we demonstrate how SILO improves sim-to-real transfer compared to standard controller tuning approaches, in addition to bringing more flexibility to controllers and reduced engineering overhead. Although evaluated on cable routing, our approach is broadly applicable to linear-deformable manipulation tasks. Our results indicate that localizing RL to deformation-critical subtasks, while handling grasping and free-space motion with classical motion primitives, is an effective strategy for scaling sim-to-real learning. Promising future directions include replacing explicit state estimation with vision-based observations using large-scale visual domain randomization, scaling to more diverse and complex harness geometries, and training cable-specific perception models to improve robustness under challenging sensing conditions.

Although our method achieves lower cycle times and higher success rates than prior approaches, several limitations remain. First, we assume that harness geometries and poses are known a priori. While this assumption holds in many industrial settings where digital models are available, it may not apply in less structured environments. In such cases, high-level policies from prior work such as hierarchical-IL \citep{luo2023multistage} could potentially be combined with our RL policies. Second, SILO currently requires synchronization between the real world and simulation, resulting in sequential start-stop motions that constrain it to quasi-static tasks. Extending SILO to more dynamic settings will require improved state synchronization and latency-aware control, which we leave to future work.

\section{Acknowledgements}
We thank our colleagues at NVIDIA for their invaluable assistance and feedback throughout this work. We extend special thanks to Hugo Hadfield for the important discussions and tooling that accelerated this research, as well as to Miles Macklin, Eric Heiden, Philipp Reist, and JC Chang for their insightful conversations on cable modeling and simulation.

\clearpage

\bibliography{example}  %

\begin{thebibliography}{58}
\providecommand{\natexlab}[1]{#1}
\providecommand{\url}[1]{\texttt{#1}}
\expandafter\ifx\csname urlstyle\endcsname\relax
  \providecommand{\doi}[1]{doi: #1}\else
  \providecommand{\doi}{doi: \begingroup \urlstyle{rm}\Url}\fi

\bibitem[Chi et~al.(2023)Chi, Feng, Du, Xu, Cousineau, Burchfiel, and Song]{chi2023diffusionpolicy}
C.~Chi, S.~Feng, Y.~Du, Z.~Xu, E.~Cousineau, B.~Burchfiel, and S.~Song.
\newblock Diffusion policy: Visuomotor policy learning via action diffusion.
\newblock In \emph{Proceedings of Robotics: Science and Systems (RSS)}, 2023.

\bibitem[Zhao et~al.(2023)Zhao, Kumar, Levine, and Finn]{DBLP:conf/rss/ZhaoKLF23}
T.~Z. Zhao, V.~Kumar, S.~Levine, and C.~Finn.
\newblock Learning fine-grained bimanual manipulation with low-cost hardware.
\newblock In K.~E. Bekris, K.~Hauser, S.~L. Herbert, and J.~Yu, editors, \emph{Robotics: Science and Systems XIX, Daegu, Republic of Korea, July 10-14, 2023}, 2023.
\newblock \doi{10.15607/RSS.2023.XIX.016}.
\newblock URL \url{https://doi.org/10.15607/RSS.2023.XIX.016}.

\bibitem[Black et~al.()Black, Brown, Driess, Esmail, Equi, Finn, Fusai, Groom, Hausman, Ichter, et~al.]{black2410pi0}
K.~Black, N.~Brown, D.~Driess, A.~Esmail, M.~Equi, C.~Finn, N.~Fusai, L.~Groom, K.~Hausman, B.~Ichter, et~al.
\newblock $\pi$0: A vision-language-action flow model for general robot control. corr, abs/2410.24164, 2024. doi: 10.48550.
\newblock \emph{arXiv preprint ARXIV.2410.24164}.

\bibitem[Kim et~al.(2025)Kim, Pertsch, Karamcheti, Xiao, Balakrishna, Nair, Rafailov, Foster, Sanketi, Vuong, Kollar, Burchfiel, Tedrake, Sadigh, Levine, Liang, and Finn]{pmlr-v270-kim25c}
M.~J. Kim, K.~Pertsch, S.~Karamcheti, T.~Xiao, A.~Balakrishna, S.~Nair, R.~Rafailov, E.~P. Foster, P.~R. Sanketi, Q.~Vuong, T.~Kollar, B.~Burchfiel, R.~Tedrake, D.~Sadigh, S.~Levine, P.~Liang, and C.~Finn.
\newblock Openvla: An open-source vision-language-action model.
\newblock In P.~Agrawal, O.~Kroemer, and W.~Burgard, editors, \emph{Proceedings of The 8th Conference on Robot Learning}, volume 270 of \emph{Proceedings of Machine Learning Research}, pages 2679--2713. PMLR, 06--09 Nov 2025.
\newblock URL \url{https://proceedings.mlr.press/v270/kim25c.html}.

\bibitem[NVIDIA et~al.(2025)NVIDIA, Bjorck, Fernando~Castañeda, Da, Ding, Fan, Fang, Fox, Hu, Huang, Jang, Jiang, Kautz, Kundalia, Lao, Li, Lin, Lin, Liu, Llontop, Magne, Mandlekar, Narayan, Nasiriany, Reed, Tan, Wang, Wang, Wang, Wang, Xiang, Xie, Xu, Xu, Ye, Yu, Zhang, Zhang, Zhao, Zheng, and Zhu]{gr00tn1_2025}
NVIDIA, J.~Bjorck, N.~C. Fernando~Castañeda, X.~Da, R.~Ding, L.~J. Fan, Y.~Fang, D.~Fox, F.~Hu, S.~Huang, J.~Jang, Z.~Jiang, J.~Kautz, K.~Kundalia, L.~Lao, Z.~Li, Z.~Lin, K.~Lin, G.~Liu, E.~Llontop, L.~Magne, A.~Mandlekar, A.~Narayan, S.~Nasiriany, S.~Reed, Y.~L. Tan, G.~Wang, Z.~Wang, J.~Wang, Q.~Wang, J.~Xiang, Y.~Xie, Y.~Xu, Z.~Xu, S.~Ye, Z.~Yu, A.~Zhang, H.~Zhang, Y.~Zhao, R.~Zheng, and Y.~Zhu.
\newblock {GR00T} {N1}: An open foundation model for generalist humanoid robots.
\newblock In \emph{ArXiv Preprint}, March 2025.

\bibitem[Luo et~al.(2023)Luo, Xu, Geng, Feng, Fang, Tan, Schaal, and Levine]{luo2023multistage}
J.~Luo, C.~Xu, X.~Geng, G.~Feng, K.~Fang, L.~Tan, S.~Schaal, and S.~Levine.
\newblock Multi-stage cable routing through hierarchical imitation learning.
\newblock \emph{arXiv pre-print}, 2023.
\newblock URL \url{https://arxiv.org/abs/2307.08927}.

\bibitem[Noseworthy et~al.(2025)Noseworthy, Tang, Wen, Handa, Kessens, Roy, Fox, Ramos, Narang, and Akinola]{noseworthy2025}
M.~Noseworthy, B.~Tang, B.~Wen, A.~Handa, C.~C. Kessens, N.~Roy, D.~Fox, F.~Ramos, Y.~Narang, and I.~Akinola.
\newblock {FORGE:} force-guided exploration for robust contact-rich manipulation under uncertainty.
\newblock \emph{{IEEE} Robotics Autom. Lett.}, 10\penalty0 (5):\penalty0 4436--4443, 2025.
\newblock \doi{10.1109/LRA.2025.3551637}.
\newblock URL \url{https://doi.org/10.1109/LRA.2025.3551637}.

\bibitem[Tang et~al.(2023)Tang, Lin, Akinola, Handa, Sukhatme, Ramos, Fox, and Narang]{tang2023}
B.~Tang, M.~A. Lin, I.~Akinola, A.~Handa, G.~S. Sukhatme, F.~Ramos, D.~Fox, and Y.~Narang.
\newblock Industreal: Transferring contact-rich assembly tasks from simulation to reality.
\newblock In K.~E. Bekris, K.~Hauser, S.~L. Herbert, and J.~Yu, editors, \emph{Robotics: Science and Systems XIX, Daegu, Republic of Korea, July 10-14, 2023}, 2023.
\newblock \doi{10.15607/RSS.2023.XIX.039}.
\newblock URL \url{https://doi.org/10.15607/RSS.2023.XIX.039}.

\bibitem[He et~al.(2025)He, Xiao, Lin, Luo, Xu, Jiang, Kautz, Liu, Shi, Wang, Fan, and Zhu]{tairanhe2025}
T.~He, W.~Xiao, T.~Lin, Z.~Luo, Z.~Xu, Z.~Jiang, J.~Kautz, C.~Liu, G.~Shi, X.~Wang, L.~J. Fan, and Y.~Zhu.
\newblock {HOVER:} versatile neural whole-body controller for humanoid robots.
\newblock In \emph{{IEEE} International Conference on Robotics and Automation, {ICRA} 2025, Atlanta, GA, USA, May 19-23, 2025}, pages 9989--9996. {IEEE}, 2025.
\newblock \doi{10.1109/ICRA55743.2025.11128549}.
\newblock URL \url{https://doi.org/10.1109/ICRA55743.2025.11128549}.

\bibitem[Rudin et~al.(2021)Rudin, Hoeller, Reist, and Hutter]{rudin2021}
N.~Rudin, D.~Hoeller, P.~Reist, and M.~Hutter.
\newblock Learning to walk in minutes using massively parallel deep reinforcement learning.
\newblock In A.~Faust, D.~Hsu, and G.~Neumann, editors, \emph{Conference on Robot Learning, 8-11 November 2021, London, {UK}}, volume 164 of \emph{Proceedings of Machine Learning Research}, pages 91--100. {PMLR}, 2021.
\newblock URL \url{https://proceedings.mlr.press/v164/rudin22a.html}.

\bibitem[Azulay et~al.(2025)Azulay, Kondap, Drake, Xie, Li, Chitta, and Goldberg]{DBLP:conf/case/AzulayKDXLCG25}
O.~Azulay, K.~Kondap, J.~Drake, S.~Xie, H.~Li, S.~Chitta, and K.~Goldberg.
\newblock {MOTORCYCLE} 1.0: Automating bimanual cable routing around fixtures on the {NIST} task board.
\newblock In \emph{21st {IEEE} International Conference on Automation Science and Engineering, {CASE} 2025, Los Angeles, CA, USA, August 17-21, 2025}, pages 2636--2641. {IEEE}, 2025.
\newblock \doi{10.1109/CASE58245.2025.11163872}.
\newblock URL \url{https://doi.org/10.1109/CASE58245.2025.11163872}.

\bibitem[Chen et~al.(2023)Chen, Bing, Wu, Meng, Kraft, Haddadin, and Knoll]{DBLP:conf/iros/0005BWMKHK23}
K.~Chen, Z.~Bing, F.~Wu, Y.~Meng, A.~Kraft, S.~Haddadin, and A.~Knoll.
\newblock Contact-aware shaping and maintenance of deformable linear objects with fixtures.
\newblock In \emph{{IROS}}, pages 1--8, 2023.
\newblock \doi{10.1109/IROS55552.2023.10341726}.
\newblock URL \url{https://doi.org/10.1109/IROS55552.2023.10341726}.

\bibitem[Jin et~al.(2022)Jin, Lian, Wang, Tomizuka, and Schaal]{DBLP:journals/ral/JinLWTS22}
S.~Jin, W.~Lian, C.~Wang, M.~Tomizuka, and S.~Schaal.
\newblock Robotic cable routing with spatial representation.
\newblock \emph{{IEEE} Robotics Autom. Lett.}, 7\penalty0 (2):\penalty0 5687--5694, 2022.
\newblock \doi{10.1109/LRA.2022.3158377}.
\newblock URL \url{https://doi.org/10.1109/LRA.2022.3158377}.

\bibitem[Waltersson et~al.(2022)Waltersson, Laezza, and Karayiannidis]{DBLP:conf/icra/WalterssonLK22}
G.~A. Waltersson, R.~Laezza, and Y.~Karayiannidis.
\newblock Planning and control for cable-routing with dual-arm robot.
\newblock In \emph{2022 International Conference on Robotics and Automation, {ICRA} 2022, Philadelphia, PA, USA, May 23-27, 2022}, pages 1046--1052. {IEEE}, 2022.
\newblock \doi{10.1109/ICRA46639.2022.9811765}.
\newblock URL \url{https://doi.org/10.1109/ICRA46639.2022.9811765}.

\bibitem[Chen et~al.(2026)Chen, Chitambar, Lam, Li, Chitta, and Goldberg]{chen2026craft}
Z.~Chen, Y.~Chitambar, T.~Lam, H.~Li, S.~Chitta, and K.~Goldberg.
\newblock Craft: Long-horizon cable routing algorithm and low-friction caging gripper.
\newblock In \emph{Proceedings of the IEEE International Conference on Robotics and Automation (ICRA)}, 2026.

\bibitem[Dong et~al.(2020)Dong, Wang, She, Sunil, Rodriguez, and Adelson]{DBLP:conf/rss/DongWSSRA20}
S.~Dong, S.~Wang, Y.~She, N.~Sunil, A.~Rodriguez, and E.~H. Adelson.
\newblock Cable manipulation with a tactile-reactive gripper.
\newblock In M.~Toussaint, A.~Bicchi, and T.~Hermans, editors, \emph{Robotics: Science and Systems XVI, Virtual Event / Corvalis, Oregon, USA, July 12-16, 2020}, 2020.
\newblock \doi{10.15607/RSS.2020.XVI.029}.
\newblock URL \url{https://doi.org/10.15607/RSS.2020.XVI.029}.

\bibitem[Li et~al.(2025)Li, Yu, Huang, Hong, and Choi]{DBLP:journals/corr/abs-2510-19268}
M.~Li, H.~Yu, Y.~Huang, Y.~Hong, and C.~Choi.
\newblock Hierarchical {DLO} routing with reinforcement learning and in-context vision-language models.
\newblock \emph{CoRR}, abs/2510.19268, 2025.
\newblock \doi{10.48550/ARXIV.2510.19268}.
\newblock URL \url{https://doi.org/10.48550/arXiv.2510.19268}.

\bibitem[Schulman et~al.(2013)Schulman, Lee, Ho, and Abbeel]{DBLP:conf/icra/SchulmanLHA13}
J.~Schulman, A.~X. Lee, J.~Ho, and P.~Abbeel.
\newblock Tracking deformable objects with point clouds.
\newblock In \emph{2013 {IEEE} International Conference on Robotics and Automation, Karlsruhe, Germany, May 6-10, 2013}, pages 1130--1137. {IEEE}, 2013.
\newblock \doi{10.1109/ICRA.2013.6630714}.
\newblock URL \url{https://doi.org/10.1109/ICRA.2013.6630714}.

\bibitem[Yan et~al.(2020)Yan, Zhu, Jin, and Bohg]{DBLP:journals/ral/YanZJB20}
M.~Yan, Y.~Zhu, N.~Jin, and J.~Bohg.
\newblock Self-supervised learning of state estimation for manipulating deformable linear objects.
\newblock \emph{{IEEE} Robotics Autom. Lett.}, 5\penalty0 (2):\penalty0 2372--2379, 2020.
\newblock \doi{10.1109/LRA.2020.2969931}.
\newblock URL \url{https://doi.org/10.1109/LRA.2020.2969931}.

\bibitem[Jin et~al.(2021)Jin, Romeres, Ragunathan, Jha, and Tomizuka]{DBLP:conf/icra/JinRRJT21}
S.~Jin, D.~Romeres, A.~Ragunathan, D.~K. Jha, and M.~Tomizuka.
\newblock Trajectory optimization for manipulation of deformable objects: Assembly of belt drive units.
\newblock In \emph{{IEEE} International Conference on Robotics and Automation, {ICRA} 2021, Xi'an, China, May 30 - June 5, 2021}, pages 10002--10008. {IEEE}, 2021.
\newblock \doi{10.1109/ICRA48506.2021.9561556}.
\newblock URL \url{https://doi.org/10.1109/ICRA48506.2021.9561556}.

\bibitem[Kim et~al.(2024)Kim, Ohmura, and Kuniyoshi]{DBLP:conf/iros/0002OK24}
H.~Kim, Y.~Ohmura, and Y.~Kuniyoshi.
\newblock Multi-task real-robot data with gaze attention for dual-arm fine manipulation.
\newblock In \emph{{IEEE/RSJ} International Conference on Intelligent Robots and Systems, {IROS} 2024, Abu Dhabi, United Arab Emirates, October 14-18, 2024}, pages 8516--8523. {IEEE}, 2024.
\newblock \doi{10.1109/IROS58592.2024.10802034}.
\newblock URL \url{https://doi.org/10.1109/IROS58592.2024.10802034}.

\bibitem[Hoang et~al.(2025)Hoang, Le, Becker, Vien, and Neumann]{DBLP:conf/iclr/HoangLBVN25}
T.~Hoang, H.~Le, P.~Becker, N.~A. Vien, and G.~Neumann.
\newblock Geometry-aware {RL} for manipulation of varying shapes and deformable objects.
\newblock In \emph{The Thirteenth International Conference on Learning Representations, {ICLR} 2025, Singapore, April 24-28, 2025}. OpenReview.net, 2025.
\newblock URL \url{https://openreview.net/forum?id=7BLXhmWvwF}.

\bibitem[Li et~al.(2022)Li, Zhang, Wong, Gokmen, Srivastava, Mart{\'{\i}}n{-}Mart{\'{\i}}n, Wang, Levine, Lingelbach, Sun, Anvari, Hwang, Sharma, Aydin, Bansal, Hunter, Kim, Lou, Matthews, Villa{-}Renteria, Tang, Tang, Xia, Savarese, Gweon, Liu, Wu, and Fei{-}Fei]{DBLP:conf/corl/0002ZWGSMWLLSAH22}
C.~Li, R.~Zhang, J.~Wong, C.~Gokmen, S.~Srivastava, R.~Mart{\'{\i}}n{-}Mart{\'{\i}}n, C.~Wang, G.~Levine, M.~Lingelbach, J.~Sun, M.~Anvari, M.~Hwang, M.~Sharma, A.~Aydin, D.~Bansal, S.~Hunter, K.~Kim, A.~Lou, C.~R. Matthews, I.~Villa{-}Renteria, J.~H. Tang, C.~Tang, F.~Xia, S.~Savarese, H.~Gweon, C.~K. Liu, J.~Wu, and L.~Fei{-}Fei.
\newblock {BEHAVIOR-1K:} {A} benchmark for embodied {AI} with 1, 000 everyday activities and realistic simulation.
\newblock In K.~Liu, D.~Kulic, and J.~Ichnowski, editors, \emph{Conference on Robot Learning, CoRL 2022, 14-18 December 2022, Auckland, New Zealand}, volume 205 of \emph{Proceedings of Machine Learning Research}, pages 80--93. {PMLR}, 2022.
\newblock URL \url{https://proceedings.mlr.press/v205/li23a.html}.

\bibitem[Lin et~al.(2020)Lin, Wang, Olkin, and Held]{DBLP:conf/corl/LinWOH20}
X.~Lin, Y.~Wang, J.~Olkin, and D.~Held.
\newblock Softgym: Benchmarking deep reinforcement learning for deformable object manipulation.
\newblock In J.~Kober, F.~Ramos, and C.~J. Tomlin, editors, \emph{4th Conference on Robot Learning, CoRL 2020, 16-18 November 2020, Virtual Event / Cambridge, MA, {USA}}, volume 155 of \emph{Proceedings of Machine Learning Research}, pages 432--448. {PMLR}, 2020.
\newblock URL \url{https://proceedings.mlr.press/v155/lin21a.html}.

\bibitem[Xing et~al.(2025)Xing, Luk, and Oh]{DBLP:conf/iclr/XingLO25}
E.~Xing, V.~Luk, and J.~Oh.
\newblock Stabilizing reinforcement learning in differentiable multiphysics simulation.
\newblock In \emph{The Thirteenth International Conference on Learning Representations, {ICLR} 2025, Singapore, April 24-28, 2025}. OpenReview.net, 2025.
\newblock URL \url{https://openreview.net/forum?id=DRiLWb8bJg}.

\bibitem[Weng et~al.(2024)Weng, Zhou, Yin, Kravberg, Varava, Navarro{-}Alarcon, and Kragic]{DBLP:journals/ral/WengZYKVNK24}
Z.~Weng, P.~Zhou, H.~Yin, A.~Kravberg, A.~Varava, D.~Navarro{-}Alarcon, and D.~Kragic.
\newblock Interactive perception for deformable object manipulation.
\newblock \emph{{IEEE} Robotics Autom. Lett.}, 9\penalty0 (9):\penalty0 7763--7770, 2024.
\newblock \doi{10.1109/LRA.2024.3431943}.
\newblock URL \url{https://doi.org/10.1109/LRA.2024.3431943}.

\bibitem[Matas et~al.(2018)Matas, James, and Davison]{DBLP:conf/corl/MatasJD18}
J.~Matas, S.~James, and A.~J. Davison.
\newblock Sim-to-real reinforcement learning for deformable object manipulation.
\newblock In \emph{2nd Annual Conference on Robot Learning, CoRL 2018, Z{\"{u}}rich, Switzerland, 29-31 October 2018, Proceedings}, volume~87 of \emph{Proceedings of Machine Learning Research}, pages 734--743. {PMLR}, 2018.
\newblock URL \url{http://proceedings.mlr.press/v87/matas18a.html}.

\bibitem[Scheikl et~al.(2023)Scheikl, Tagliabue, Gyenes, Wagner, Dall'Alba, Fiorini, and Mathis{-}Ullrich]{DBLP:journals/ral/ScheiklTGWDFM23}
P.~M. Scheikl, E.~Tagliabue, B.~Gyenes, M.~Wagner, D.~Dall'Alba, P.~Fiorini, and F.~Mathis{-}Ullrich.
\newblock Sim-to-real transfer for visual reinforcement learning of deformable object manipulation for robot-assisted surgery.
\newblock \emph{{IEEE} Robotics Autom. Lett.}, 8\penalty0 (2):\penalty0 560--567, 2023.
\newblock \doi{10.1109/LRA.2022.3227873}.
\newblock URL \url{https://doi.org/10.1109/LRA.2022.3227873}.

\bibitem[Wu et~al.(2020)Wu, Yan, Kurutach, Pinto, and Abbeel]{DBLP:conf/rss/WuYKPA20}
Y.~Wu, W.~Yan, T.~Kurutach, L.~Pinto, and P.~Abbeel.
\newblock Learning to manipulate deformable objects without demonstrations.
\newblock In M.~Toussaint, A.~Bicchi, and T.~Hermans, editors, \emph{Robotics: Science and Systems XVI, Virtual Event / Corvalis, Oregon, USA, July 12-16, 2020}, 2020.
\newblock \doi{10.15607/RSS.2020.XVI.065}.
\newblock URL \url{https://doi.org/10.15607/RSS.2020.XVI.065}.

\bibitem[Coumans and Bai(2016--2019)]{coumans2019}
E.~Coumans and Y.~Bai.
\newblock Pybullet, a python module for physics simulation for games, robotics and machine learning.
\newblock \url{http://pybullet.org}, 2016--2019.

\bibitem[{NVIDIA}()]{NVIDIA_Isaac_Sim}
{NVIDIA}.
\newblock {Isaac Sim}.
\newblock URL \url{https://github.com/isaac-sim/IsaacSim}.

\bibitem[Todorov et~al.(2012)Todorov, Erez, and Tassa]{todorov2012mujoco}
E.~Todorov, T.~Erez, and Y.~Tassa.
\newblock Mujoco: A physics engine for model-based control.
\newblock In \emph{2012 IEEE/RSJ International Conference on Intelligent Robots and Systems}, pages 5026--5033. IEEE, 2012.
\newblock \doi{10.1109/IROS.2012.6386109}.

\bibitem[{Newton Contributors}(2025)]{newton}
{Newton Contributors}.
\newblock {Newton}: {GPU}-accelerated physics simulation for robotics, and simulation research., 2025.
\newblock URL \url{https://github.com/newton-physics/newton}.

\bibitem[Xiang et~al.(2020)Xiang, Qin, Mo, Xia, Zhu, Liu, Liu, Jiang, Yuan, Wang, Yi, Chang, Guibas, and Su]{DBLP:conf/cvpr/XiangQMXZLLJYWY20}
F.~Xiang, Y.~Qin, K.~Mo, Y.~Xia, H.~Zhu, F.~Liu, M.~Liu, H.~Jiang, Y.~Yuan, H.~Wang, L.~Yi, A.~X. Chang, L.~J. Guibas, and H.~Su.
\newblock {SAPIEN:} {A} simulated part-based interactive environment.
\newblock In \emph{2020 {IEEE/CVF} Conference on Computer Vision and Pattern Recognition, {CVPR} 2020, Seattle, WA, USA, June 13-19, 2020}, pages 11094--11104. Computer Vision Foundation / {IEEE}, 2020.
\newblock \doi{10.1109/CVPR42600.2020.01111}.
\newblock URL \url{https://openaccess.thecvf.com/content\_CVPR\_2020/html/Xiang\_SAPIEN\_A\_SimulAted\_Part-Based\_Interactive\_ENvironment\_CVPR\_2020\_paper.html}.

\bibitem[Gu et~al.(2023)Gu, Xiang, Li, Ling, Liu, Mu, Tang, Tao, Wei, Yao, Yuan, Xie, Huang, Chen, and Su]{gu2023maniskill2}
J.~Gu, F.~Xiang, X.~Li, Z.~Ling, X.~Liu, T.~Mu, Y.~Tang, S.~Tao, X.~Wei, Y.~Yao, X.~Yuan, P.~Xie, Z.~Huang, R.~Chen, and H.~Su.
\newblock Maniskill2: A unified benchmark for generalizable manipulation skills.
\newblock In \emph{International Conference on Learning Representations}, 2023.

\bibitem[Tao et~al.(2025)Tao, Xiang, Shukla, Qin, Hinrichsen, Yuan, Bao, Lin, Liu, kai Chan, Gao, Li, Mu, Xiao, Gurha, Rajesh, Choi, Chen, Huang, Calandra, Chen, Luo, and Su]{taomaniskill3}
S.~Tao, F.~Xiang, A.~Shukla, Y.~Qin, X.~Hinrichsen, X.~Yuan, C.~Bao, X.~Lin, Y.~Liu, T.~kai Chan, Y.~Gao, X.~Li, T.~Mu, N.~Xiao, A.~Gurha, V.~N. Rajesh, Y.~W. Choi, Y.-R. Chen, Z.~Huang, R.~Calandra, R.~Chen, S.~Luo, and H.~Su.
\newblock Maniskill3: Gpu parallelized robotics simulation and rendering for generalizable embodied ai.
\newblock \emph{Robotics: Science and Systems}, 2025.

\bibitem[Macklin(2022)]{warp2022}
M.~Macklin.
\newblock Warp: A high-performance python framework for gpu simulation and graphics.
\newblock \url{https://github.com/nvidia/warp}, March 2022.
\newblock NVIDIA GPU Technology Conference (GTC).

\bibitem[Huang et~al.(2021)Huang, Hu, Du, Zhou, Su, Tenenbaum, and Gan]{DBLP:conf/iclr/HuangHDZ0TG21}
Z.~Huang, Y.~Hu, T.~Du, S.~Zhou, H.~Su, J.~B. Tenenbaum, and C.~Gan.
\newblock Plasticinelab: {A} soft-body manipulation benchmark with differentiable physics.
\newblock In \emph{9th International Conference on Learning Representations, {ICLR} 2021, Virtual Event, Austria, May 3-7, 2021}. OpenReview.net, 2021.
\newblock URL \url{https://openreview.net/forum?id=xCcdBRQEDW}.

\bibitem[Hu et~al.(2019)Hu, Li, Anderson, Ragan-Kelley, and Durand]{hu2019taichi}
Y.~Hu, T.-M. Li, L.~Anderson, J.~Ragan-Kelley, and F.~Durand.
\newblock Taichi: a language for high-performance computation on spatially sparse data structures.
\newblock \emph{ACM Transactions on Graphics (TOG)}, 38\penalty0 (6):\penalty0 201, 2019.

\bibitem[Li et~al.(2023)Li, Antonova, Sadigh, and Bohg]{DBLP:conf/icra/LiASB23}
M.~Li, R.~Antonova, D.~Sadigh, and J.~Bohg.
\newblock Learning tool morphology for contact-rich manipulation tasks with differentiable simulation.
\newblock In \emph{{IEEE} International Conference on Robotics and Automation, {ICRA} 2023, London, UK, May 29 - June 2, 2023}, pages 1859--1865. {IEEE}, 2023.
\newblock \doi{10.1109/ICRA48891.2023.10161453}.
\newblock URL \url{https://doi.org/10.1109/ICRA48891.2023.10161453}.

\bibitem[Xu et~al.(2021)Xu, Chen, Zlokapa, Foshey, Matusik, Sueda, and Agrawal]{xu2021diffhand}
J.~Xu, T.~Chen, L.~Zlokapa, M.~Foshey, W.~Matusik, S.~Sueda, and P.~Agrawal.
\newblock {An End-to-End Differentiable Framework for Contact-Aware Robot Design}.
\newblock In \emph{Proceedings of Robotics: Science and Systems}, Virtual, July 2021.
\newblock \doi{10.15607/RSS.2021.XVII.008}.

\bibitem[Jacob et~al.(2024)Jacob, Bandyopadhyay, Williams, Borges, and Ramos]{DBLP:journals/ral/JacobBWBR24}
J.~Jacob, T.~Bandyopadhyay, J.~Williams, P.~V.~K. Borges, and F.~Ramos.
\newblock Learning to simulate tree-branch dynamics for manipulation.
\newblock \emph{{IEEE} Robotics Autom. Lett.}, 9\penalty0 (2):\penalty0 1748--1755, 2024.
\newblock \doi{10.1109/LRA.2024.3349830}.
\newblock URL \url{https://doi.org/10.1109/LRA.2024.3349830}.

\bibitem[Makoviychuk et~al.(2021)Makoviychuk, Wawrzyniak, Guo, Lu, Storey, Macklin, Hoeller, Rudin, Allshire, Handa, and State]{makoviychuk2021isaac}
V.~Makoviychuk, L.~Wawrzyniak, Y.~Guo, M.~Lu, K.~Storey, M.~Macklin, D.~Hoeller, N.~Rudin, A.~Allshire, A.~Handa, and G.~State.
\newblock Isaac gym: High performance gpu-based physics simulation for robot learning, 2021.

\bibitem[Jiang et~al.(2024)Jiang, Wang, Zhang, Wu, and Fei-Fei]{jiang2024transic}
Y.~Jiang, C.~Wang, R.~Zhang, J.~Wu, and L.~Fei-Fei.
\newblock Transic: Sim-to-real policy transfer by learning from online correction.
\newblock In \emph{Conference on Robot Learning}, 2024.

\bibitem[Christen et~al.(2023)Christen, Yang, P\'{e}rez-D'Arpino, Hilliges, Fox, and Chao]{christen:cvpr2023}
S.~Christen, W.~Yang, C.~P\'{e}rez-D'Arpino, O.~Hilliges, D.~Fox, and Y.-W. Chao.
\newblock Learning human-to-robot handovers from point clouds.
\newblock In \emph{IEEE/CVF Conference on Computer Vision and Pattern Recognition (CVPR)}, 2023.

\bibitem[Qin et~al.(2022)Qin, Huang, Yin, Su, and Wang]{DBLP:conf/corl/QinHY0022}
Y.~Qin, B.~Huang, Z.~Yin, H.~Su, and X.~Wang.
\newblock Dexpoint: Generalizable point cloud reinforcement learning for sim-to-real dexterous manipulation.
\newblock In K.~Liu, D.~Kulic, and J.~Ichnowski, editors, \emph{Conference on Robot Learning, CoRL 2022, 14-18 December 2022, Auckland, New Zealand}, volume 205 of \emph{Proceedings of Machine Learning Research}, pages 594--605. {PMLR}, 2022.
\newblock URL \url{https://proceedings.mlr.press/v205/qin23a.html}.

\bibitem[Tobin et~al.(2017)Tobin, Fong, Ray, Schneider, Zaremba, and Abbeel]{DBLP:conf/iros/TobinFRSZA17}
J.~Tobin, R.~Fong, A.~Ray, J.~Schneider, W.~Zaremba, and P.~Abbeel.
\newblock Domain randomization for transferring deep neural networks from simulation to the real world.
\newblock In \emph{2017 {IEEE/RSJ} International Conference on Intelligent Robots and Systems, {IROS} 2017, Vancouver, BC, Canada, September 24-28, 2017}, pages 23--30. {IEEE}, 2017.
\newblock \doi{10.1109/IROS.2017.8202133}.
\newblock URL \url{https://doi.org/10.1109/IROS.2017.8202133}.

\bibitem[Singh et~al.(2024)Singh, Allshire, Handa, Ratliff, and Wyk]{DBLP:journals/corr/abs-2412-01791}
R.~Singh, A.~Allshire, A.~Handa, N.~D. Ratliff, and K.~V. Wyk.
\newblock Dextrah-rgb: Visuomotor policies to grasp anything with dexterous hands.
\newblock \emph{CoRR}, abs/2412.01791, 2024.
\newblock \doi{10.48550/ARXIV.2412.01791}.
\newblock URL \url{https://doi.org/10.48550/arXiv.2412.01791}.

\bibitem[Zakka et~al.(2025)Zakka, Tabanpour, Liao, Haiderbhai, Holt, Luo, Allshire, Frey, Sreenath, Kahrs, Sferrazza, Tassa, and Abbeel]{mujoco_playground_2025}
K.~Zakka, B.~Tabanpour, Q.~Liao, M.~Haiderbhai, S.~Holt, J.~Y. Luo, A.~Allshire, E.~Frey, K.~Sreenath, L.~A. Kahrs, C.~Sferrazza, Y.~Tassa, and P.~Abbeel.
\newblock Mujoco playground: An open-source framework for gpu-accelerated robot learning and sim-to-real transfer., 2025.
\newblock URL \url{https://github.com/google-deepmind/mujoco_playground}.

\bibitem[Mittal et~al.(2025)Mittal, Roth, Tigue, Richard, Zhang, Du, Serrano-Muñoz, Yao, Zurbrügg, Rudin, Wawrzyniak, Rakhsha, Denzler, Heiden, Borovicka, Ahmed, Akinola, Anwar, Carlson, Feng, Garg, Gasoto, Gulich, Guo, Gussert, Hansen, Kulkarni, Li, Liu, Makoviychuk, Malczyk, Mazhar, Moghani, Murali, Noseworthy, Poddubny, Ratliff, Rehberg, Schwarke, Singh, Smith, Tang, Thaker, Trepte, Wyk, Yu, Millane, Ramasamy, Steiner, Subramanian, Volk, Chen, Jawale, Kuruttukulam, Lin, Mandlekar, Patzwaldt, Welsh, Zhao, Anes, Lafleche, Moënne-Loccoz, Park, Stepinski, Gelder, Amevor, Carius, Chang, Chen, de~Heras~Ciechomski, Daviet, Mohajerani, von Muralt, Reutskyy, Sauter, Schirm, Shi, Terdiman, Vilella, Widmer, Yeoman, Chen, Grizan, Li, Li, Smith, Wiltz, Alexis, Chang, Chu, Fan, Farshidian, Handa, Huang, Hutter, Narang, Pouya, Sheng, Zhu, Macklin, Moravanszky, Reist, Guo, Hoeller, and State]{mittal2025isaaclab}
M.~Mittal, P.~Roth, J.~Tigue, A.~Richard, O.~Zhang, P.~Du, A.~Serrano-Muñoz, X.~Yao, R.~Zurbrügg, N.~Rudin, L.~Wawrzyniak, M.~Rakhsha, A.~Denzler, E.~Heiden, A.~Borovicka, O.~Ahmed, I.~Akinola, A.~Anwar, M.~T. Carlson, J.~Y. Feng, A.~Garg, R.~Gasoto, L.~Gulich, Y.~Guo, M.~Gussert, A.~Hansen, M.~Kulkarni, C.~Li, W.~Liu, V.~Makoviychuk, G.~Malczyk, H.~Mazhar, M.~Moghani, A.~Murali, M.~Noseworthy, A.~Poddubny, N.~Ratliff, W.~Rehberg, C.~Schwarke, R.~Singh, J.~L. Smith, B.~Tang, R.~Thaker, M.~Trepte, K.~V. Wyk, F.~Yu, A.~Millane, V.~Ramasamy, R.~Steiner, S.~Subramanian, C.~Volk, C.~Chen, N.~Jawale, A.~V. Kuruttukulam, M.~A. Lin, A.~Mandlekar, K.~Patzwaldt, J.~Welsh, H.~Zhao, F.~Anes, J.-F. Lafleche, N.~Moënne-Loccoz, S.~Park, R.~Stepinski, D.~V. Gelder, C.~Amevor, J.~Carius, J.~Chang, A.~H. Chen, P.~de~Heras~Ciechomski, G.~Daviet, M.~Mohajerani, J.~von Muralt, V.~Reutskyy, M.~Sauter, S.~Schirm, E.~L. Shi, P.~Terdiman, K.~Vilella, T.~Widmer, G.~Yeoman, T.~Chen, S.~Grizan, C.~Li, L.~Li, C.~Smith, R.~Wiltz,
  K.~Alexis, Y.~Chang, D.~Chu, L.~J. Fan, F.~Farshidian, A.~Handa, S.~Huang, M.~Hutter, Y.~Narang, S.~Pouya, S.~Sheng, Y.~Zhu, M.~Macklin, A.~Moravanszky, P.~Reist, Y.~Guo, D.~Hoeller, and G.~State.
\newblock Isaac lab: A gpu-accelerated simulation framework for multi-modal robot learning.
\newblock \emph{arXiv preprint arXiv:2511.04831}, 2025.
\newblock URL \url{https://arxiv.org/abs/2511.04831}.

\bibitem[Schulman et~al.(2017)Schulman, Wolski, Dhariwal, Radford, and Klimov]{schulman-ppo}
J.~Schulman, F.~Wolski, P.~Dhariwal, A.~Radford, and O.~Klimov.
\newblock Proximal policy optimization algorithms.
\newblock \emph{CoRR}, abs/1707.06347, 2017.
\newblock URL \url{http://arxiv.org/abs/1707.06347}.

\bibitem[Wen et~al.(2025)Wen, Trepte, Aribido, Kautz, Gallo, and Birchfield]{wen2025stereo}
B.~Wen, M.~Trepte, J.~Aribido, J.~Kautz, O.~Gallo, and S.~Birchfield.
\newblock Foundationstereo: Zero-shot stereo matching.
\newblock \emph{arXiv}, 2025.

\bibitem[Ravi et~al.(2025)Ravi, Gabeur, Hu, Hu, Ryali, Ma, Khedr, R{\"a}dle, Rolland, Gustafson, Mintun, Pan, Alwala, Carion, Wu, Girshick, Dollar, and Feichtenhofer]{ravi2025sam}
N.~Ravi, V.~Gabeur, Y.-T. Hu, R.~Hu, C.~Ryali, T.~Ma, H.~Khedr, R.~R{\"a}dle, C.~Rolland, L.~Gustafson, E.~Mintun, J.~Pan, K.~V. Alwala, N.~Carion, C.-Y. Wu, R.~Girshick, P.~Dollar, and C.~Feichtenhofer.
\newblock {SAM} 2: Segment anything in images and videos.
\newblock In \emph{The Thirteenth International Conference on Learning Representations}, 2025.
\newblock URL \url{https://openreview.net/forum?id=Ha6RTeWMd0}.

\bibitem[Wen et~al.(2025)Wen, Dewan, and Birchfield]{wen2025fast}
B.~Wen, S.~Dewan, and S.~Birchfield.
\newblock Fast-foundationstereo: Real-time zero-shot stereo matching.
\newblock \emph{arXiv preprint arXiv:2512.11130}, 2025.

\bibitem[Carion et~al.(2025)Carion, Gustafson, Hu, Debnath, Hu, Suris, Ryali, Alwala, Khedr, Huang, et~al.]{carion2025sam}
N.~Carion, L.~Gustafson, Y.-T. Hu, S.~Debnath, R.~Hu, D.~Suris, C.~Ryali, K.~V. Alwala, H.~Khedr, A.~Huang, et~al.
\newblock Sam 3: Segment anything with concepts.
\newblock \emph{arXiv preprint arXiv:2511.16719}, 2025.

\bibitem[Schneider()]{Schneider_franky_High-Level_Control}
T.~Schneider.
\newblock {franky: High-Level Control Library for Franka Robots}.
\newblock URL \url{https://github.com/TimSchneider42/franky}.

\bibitem[Berscheid and Kr{\"o}ger(2021)]{berscheid2021jerk}
L.~Berscheid and T.~Kr{\"o}ger.
\newblock Jerk-limited real-time trajectory generation with arbitrary target states.
\newblock \emph{Robotics: Science and Systems XVII}, 2021.

\bibitem[Lin et~al.(2025)Lin, Sachdev, Fan, Malik, and Zhu]{lin2025sim}
T.~Lin, K.~Sachdev, L.~Fan, J.~Malik, and Y.~Zhu.
\newblock Sim-to-real reinforcement learning for vision-based dexterous manipulation on humanoids.
\newblock \emph{arXiv:2502.20396}, 2025.

\end{thebibliography}
\newpage
\appendix
For videos, see the supplemental video / our website: \href{https://silo-cable-routing.github.io/}{https://silo-cable-routing.github.io/}
\section{State Estimation System}
\label{appendix:state_estimation}

\begin{figure*}[h]
    \centering
    \includegraphics[width=\linewidth]{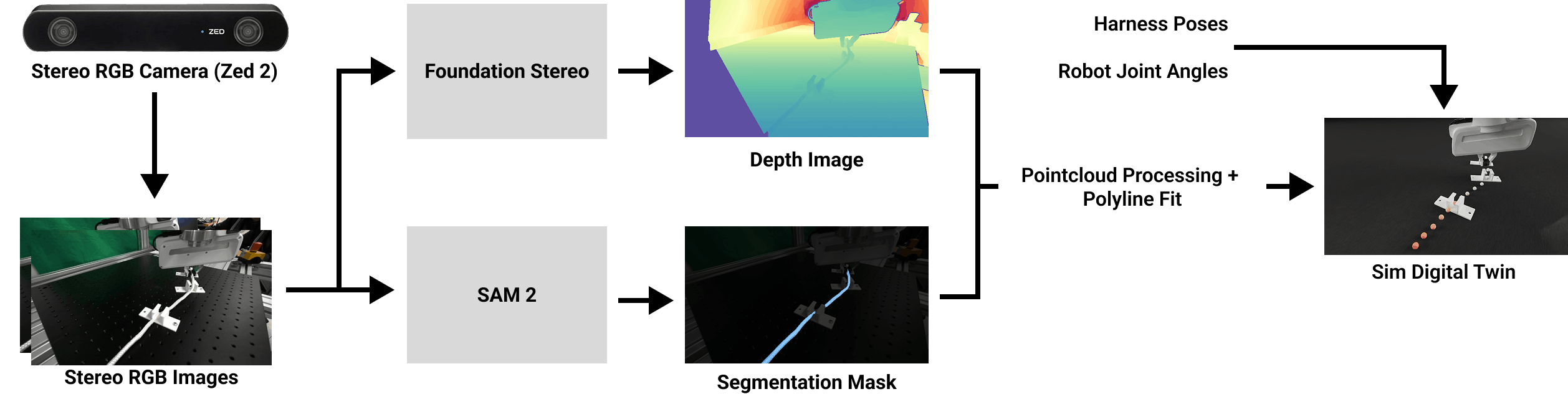}
    \caption{Visual overview of the state estimation system used to build sim digital twins during deployment. In the pointcloud processing step we remove outlier points by only keeping points where in a given inlier radius there are a minimum number of other points.}
    \label{fig:state_estimation_system}
\end{figure*}

Figure \ref{fig:state_estimation_system} shows the pipeline used to predict cable points and construct the sim digital twin during deployment. In subsequent subsections, we discuss implementation details for each component. The exact choices of hyperparameters for the tunable components of the state estimation system are shown in Table \ref{table:state_estimation_parameters}.

\subsection{Foundation Stereo: Depth Image Prediction}
Two RGB images are passed to the Foundation Stereo \citep{wen2025stereo} model which initially predicts a disparity map. Using the left stereo camera intrinsics and the right stereo camera's extrinsics relative to the left stereo camera, a depth image is generated from the perspective of the left stereo camera. In practice, we have observed that a typical failure mode of Foundation Stereo with stereo RGB cameras is when objects are too close to the camera. We found the Zed 2 camera to work the best, while other cameras like the Zed mini work if placed at a far enough distance.

\subsection{SAM2: Segmentation Prediction}
We use SAM2 \citep{ravi2025sam} for segmentation. SAM2 is only given the left RGB image to process. Each image segmentation is conditioned on the previous image's segmentation during deployment. The first image is semi-automatically segmented by having the user click points (usually about 1 to 3 points are sufficient) to indicate which parts of the first frame are part of the cable. The initial annotation is reused for the same cable across all experiments. In this paper, we primarily tested on 4 different cables so only 4 annotations were needed.

After SAM2 generates a segmentation mask, where pixels corresponding to the cable are set to 1 and other pixels set to 0, we perform a distance transform on the mask via the OpenCV library. The distance transform returns an image with the Euclidean distance of each non-zero pixel to the nearest zero pixel. We shrink the mask by remove masked pixels where said distance is greater than a given segmentation margin parameter (we use 1.0).

\subsection{Segmented Cable Pointcloud}
We first apply the segmentation mask to the depth image, isolating the depth values of the pixels corresponding to the cable. Using the camera intrinsics and extrinsics (calibrated to the world frame centered at the robot's base), the remaining pixels are lifted into a 3D pointcloud in the world frame. We then perform outlier removal, by only keeping the points that have a minimum number of neighbors (6) within a specified inlier radius (0.004m).

\begin{algorithm*}[t]
\caption{Polyline Fitting (Equal-Length Polyline from Cable Pointcloud)}
\label{alg:polyline-fit}
\begin{algorithmic}[1]
\Require Cable pointcloud $P=\{p_1,\ldots,p_n\}\subset\mathbb{R}^3$, decimation $d$ (number of polyline points), target cable length $l$, bin radius factor $r_{\text{poly}}$
\Ensure Equal-length ordered polyline $Q=\{q_1,\ldots,q_d\}\subset\mathbb{R}^3$

\State $Q \gets \{q_1,\ldots,q_d\}$ \Comment{$q_i$ are the output polyline vertices (ordered from one end to the other)}
\Statex
\State \Comment{\textbf{Stage 1: Project points onto dominant direction}}
\State $\mu \gets \frac{1}{n}\sum_{i=1}^n p_i$ \Comment{Cable pointcloud centroid}

\State $X \gets [p_1-\mu;\;\ldots;\;p_n-\mu] \in \mathbb{R}^{n\times 3}$
\State $[U,S,V^\top] \gets \mathrm{SVD}(X)$ \Comment Run Singular Value Decomposition on the centered cable pointcloud
\State $\sigma \gets V_{:,1}$ \Comment{Principal direction (first right-singular vector)}

\State $x_i \gets (p_i-\mu)^\top \sigma \;\;\;\; \forall i$ \Comment{Project points onto principal axis to obtain a 1D coordinate}
\State $x_{\min} \gets \min_i x_i,\;\; x_{\max} \gets \max_i x_i$
\State $\bar{x}_i \gets \frac{x_i-x_{\min}}{x_{\max}-x_{\min}} \in [0,1] \;\;\;\; \forall i$ \Comment{Normalize projected points into normalized coordinates $[0,1]$}

\Statex
\State \Comment{\textbf{Stage 2: Bin projected points to obtain an ordered initial polyline}}
\For{$i=1$ to $d$}
    \State $t \gets \frac{i-1}{d-1}$ \Comment{Bin center in normalized coordinate}
    \State $\epsilon \gets \frac{r_{\text{poly}}}{d-1}$ \Comment{Bin half-width (scale-aware); $r_{\text{poly}}$ is a constant (e.g., $0.2$)}
    \State $I_i \gets \{j \mid |\bar{x}_j - t| < \epsilon, x_j\}$ \Comment{Indices of points whose 1D coordinate falls near bin center}
    \State $q_i \gets \frac{1}{|I_i|}\sum_{j\in I_i} p_j$ \Comment{Average the 3D points assigned to this bin}
\EndFor

\Statex
\State \Comment{\textbf{Stage 3: Resample the polyline to enforce an equal length polyline}}
\State $r_1 \gets 0$
\For{$k=2$ to $d$}
    \State $r_k \gets r_{k-1} + \|q_k - q_{k-1}\|_2$ \Comment{Cumulative distances along the initial polyline}
\EndFor

\State $q_1 \gets q_1$ \Comment{Anchor the first point}
\For{$i=2$ to $d$}
    \State $t \gets \frac{i-1}{d-1}\cdot l$ \Comment{Target length from $q_1$ to $q_i$ for an equal-length polyline}
    \State $j \gets \max\{k \mid r_k \le t\}$ \Comment{Segment index such that $t$ lies between $r_j$ and $r_{j+1}$}
    \State $\Delta \gets r_{j+1}-r_j$
        \State $\alpha \gets \frac{t-r_j}{\Delta} \in [0,1]$
        \State $q_i \gets q_j + \alpha\,(q_{j+1}-q_j)$ \Comment{Linear interpolation to update $q_i$ to achieve target length $t$}
\EndFor

\State \Return $Q$
\end{algorithmic}
\end{algorithm*}

To obtain a sim-to-real compatible representation, we fit an ordered, equal-length polyline to the segmented cable point cloud. This representation matches the cable discretization used in simulation and provides a compact description of local cable geometry near the gripper.
The polyline fitting procedure is described in Algorithm \ref{alg:polyline-fit} and proceeds as follows:
\begin{enumerate}
    \item Cable Pointcloud Projection:
 We compute the dominant direction $\sigma$ of the point cloud using singular value decomposition (SVD) and project all points onto this axis. This provides a 1D ordering parameter along the cable length.
    \item Binning:
 The projected points are normalized to $[0,1]\times[0, 1]\times[0,1]$ and partitioned into $d$ evenly spaced bins, where $d$ is the cable decimation. Points falling within each bin are averaged to produce an initial ordered polyline.
If no points fall into a bin (e.g., due to occlusion), the closest valid polyline point is reused to maintain continuity.
    \item Resampling:
 The polyline is resampled to enforce equal spacing between adjacent points, ensuring compatibility with the fixed-length link representation used in simulation.
\end{enumerate}
This procedure is robust to moderate occlusions (e.g., partial harness blockage), as disconnected visible segments are implicitly reconnected during polyline construction. The method is fast to run but trades off generalization to some more ``twisted'' cable shapes as it does not work well when the cable curves into large U shapes or has crossovers. For our test cases the polyline fitting algorithm works sufficiently well. One reason is because our observations only include the closest 4 cable points to the gripper and thus does not need an accurate prediction of the entire cable.

\subsection{Sim Digital Twin}
We can read ground truth joint angles from the robot, and harness poses and shapes are provided a priori. Along with the calibrated camera extrinsics and intrinsics, we reconstruct the robot, harnesses and the real world camera in simulation accordingly. The predicted cable points after polyline fitting are then spawned in as visual spheres to show where the predicted cable points are. The color of the spheres shows how the points are ordered, which naturally arises from the polyline fitting procedure.

\subsection{Related Work}
\cite{DBLP:conf/case/AzulayKDXLCG25} uses a state estimation system for cables but takes a different approach. Their approach can handle more turns and handle some crossovers in cables compared to our system, but otherwise is limited to overhead cameras and planar cables. In contrast, our system predicts the 3D positions of points on the cable and is thus not limited to overhead cameras or planar cables.

\begin{table}
\small
\centering
\begin{tabular}{ll}
\hline
\rowcolor{gray!30}
\textbf{Hyperparameter} & \textbf{Value} \\
\hline
\rowcolor{gray!10}
\textbf{Cameras} & \\
\hline
Camera Type & Zed 2 \\
Camera Resolution & 640$\times$ 360\\
\hline
\rowcolor{gray!10}
\textbf{Foundation Stereo and Depth Estimation} & \\
\hline
Model checkpoint & 23-51-11\\
Valid iterations & 4 \\
Inlier Radius & 0.004m \\
Inlier Count & 6\\
\hline
\rowcolor{gray!10}
\textbf{SAM 2 and Segmentation} & \\
\hline
Model checkpoint & sam2.1\_hiera\_tiny \\
Segmentation Margin & 1.0 \\
\hline
\rowcolor{gray!10}
\textbf{Polyline Fitting} & \\
\hline
$r_\text{polyline}$ & 0.2 \\
\hline
\end{tabular}
\caption{Hyperparameters for state-estimation pipeline}
\label{table:state_estimation_parameters}
\end{table}

\section{Motion Primitives}
\label{appendix:motion_primitives}
\begin{figure}[h]
    \centering
    \includegraphics[width=1\linewidth]{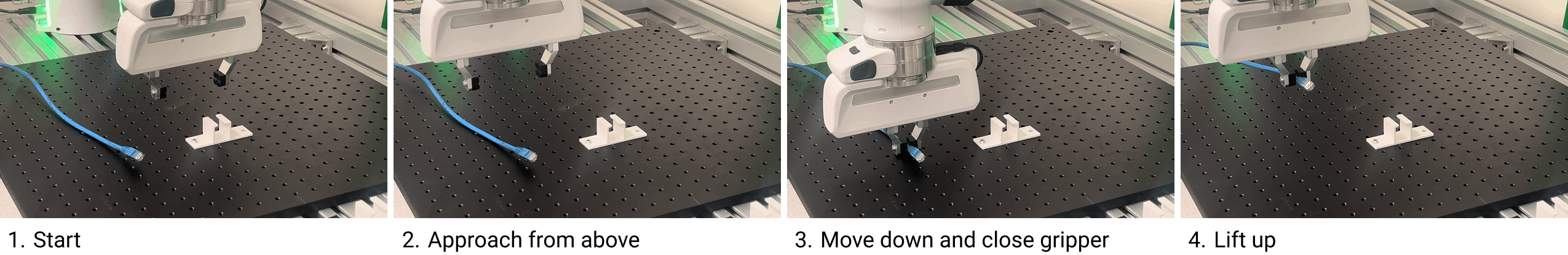}
    \caption{The sequence of movements for the \textit{GraspCable} primitive in the real world.}
    \label{fig:grasp_cable_primitive_real}
\end{figure}
\begin{figure}[h]
    \centering
    \includegraphics[width=1\linewidth]{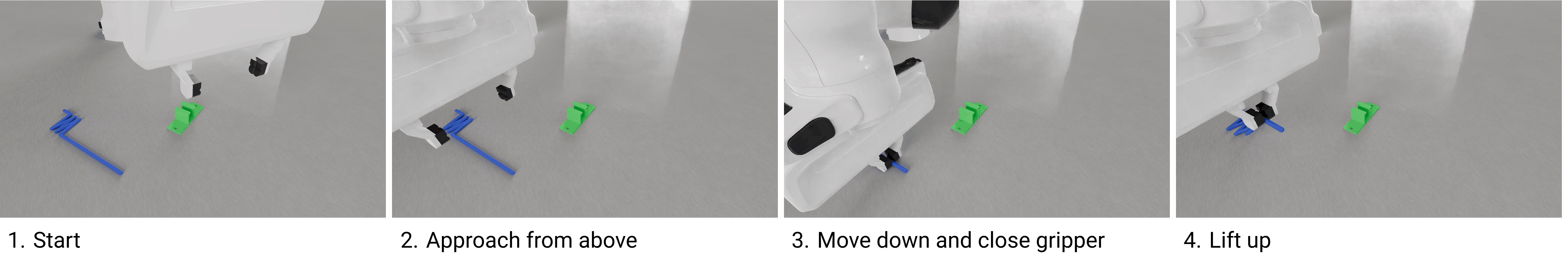}
    \caption{The sequence of movements for the \textit{GraspCable} primitive in simulation.}
    \label{fig:grasp_cable_primitive_sim}
\end{figure}
\subsection{\textit{GraspCable}}
Visualization of the \textit{GraspCable} primitive deployed in the real world is shown in Figure \ref{fig:grasp_cable_primitive_real}. The same primitive when run in simulation is shown in Figure \ref{fig:grasp_cable_primitive_sim}.

\subsection{\textit{MoveToHarness}}
\begin{figure*}
    \centering
    \includegraphics[width=1\linewidth]{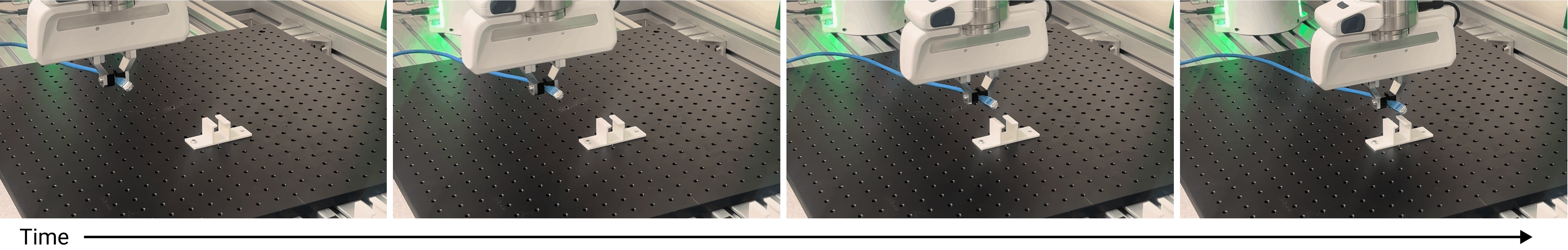}
    \caption{The sequence of movements for the \textit{MoveToHarness} primitive in the real world.}
    \label{fig:move_to_harness_primitive_real}
\end{figure*}

\begin{figure*}
    \centering
    \includegraphics[width=1\linewidth]{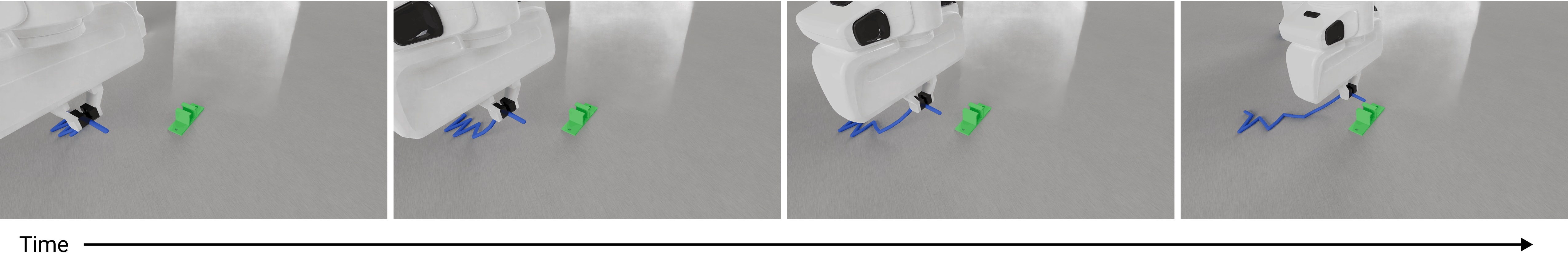}
    \caption{The sequence of movements for the \textit{MoveToHarness} primitive in simulation.}
    \label{fig:move_to_harness_primitive_sim}
\end{figure*}

Visualization of the \textit{MoveToHarness} primitive deployed in the real world is shown in Figure \ref{fig:move_to_harness_primitive_real}. The same primitive when run in simulation is shown in Figure \ref{fig:move_to_harness_primitive_sim}.

\section{SILO: Sim In the Loop}
\label{appendix:silo}
This section provides additional technical details and motivation for Simulation-in-the-Loop (SILO), the deployment framework used to bridge the sim-to-real gap for cable routing. SILO reuses the simulator’s internal dynamics during real-world execution to mitigate sim-to-real mismatch, enforce collision safety, and reduce system engineering overhead.

\subsection{Overview and Assumptions}
SILO operates by tightly synchronizing a simulated digital twin with the physical robot during deployment. Rather than directly executing policy actions on the real robot, actions are first applied in simulation, and the resulting simulated joint targets are then tracked by the physical robot.

The following assumptions underlie SILO:
\begin{itemize}
    \item \textbf{Quasi-static manipulation}: Robot and object motion primarily occur during commanded actions; unmodeled dynamics between control steps are negligible.
    \item \textbf{Accurate robot kinematics}: The simulator’s kinematic model of the robot matches the real robot within calibration tolerance.
    \item \textbf{Known static geometry}: Harness geometry and poses are known and fixed during execution.
\end{itemize}
SILO is not designed for highly dynamic manipulation (e.g., throwing objects), where tight real-time coupling between physics and control would be required. However, these assumptions are reasonable for many assembly-type problems like cable routing.

\subsection{Sim-to-real Dynamics and Controller Mismatch}
Sim-to-real transfer is limited by several sources of mismatch between simulation and the physical world. Three primary contributors include:

\begin{itemize}
    \item \textbf{Approximate dynamics models}: Physics engines necessarily approximate real-world dynamics, particularly for contact-rich and deformable objects such as cables. Even with careful parameter tuning, simulated dynamics remains an abstraction of physical behavior.
    \item \textbf{Numerical solver residuals}: Articulated dynamics, joint constraints, and contacts are solved iteratively and only up to finite numerical tolerance. Residual errors accumulate over time and vary with solver settings, timestep, and backend implementation.
    \item \textbf{Controller dynamics mismatch}: Low-level controllers in simulation only partially approximate those used on real robots. This mismatch is especially pronounced in GPU-parallelized simulators, where controllers are optimized for throughput rather than exact convergence within a single physics step.
\end{itemize}

Even with a perfectly accurate dynamics model, solver residuals and controller discrepancies would still introduce systematic divergence between simulated and real trajectories.
For example, in PhysX-based GPU simulations, a commanded joint target typically is not reached within a single physics step; the realized motion also depends on the current physics solver configurations. Full convergence where the commanded joint targets are achieved often require multiple physics steps. During RL training, policies implicitly adapt to these simulator-specific behaviors. Directly executing the same actions on a real robot, whose controller dynamics differ, can therefore lead to accumulated error and degraded performance.

SILO mitigates these sources of sim-to-real gap by reusing the simulator’s internal solver and controller dynamics during deployment. Rather than transferring actions directly to the real robot, SILO transfers simulated outcomes, treating the simulator as the reference for motion evolution. This approach effectively bypasses controller mismatch and solver residual discrepancies without requiring explicit system identification or manual controller tuning.

\subsection{Collision Avoidance}
An important byproduct of SILO is inherent collision avoidance. Since actions are first executed in a simulated digital twin that includes the harness geometry, joint configurations that would result in penetration generally cannot occur at any given physics step.
This allows collision constraints to be enforced at deployment time without explicit collision penalties during RL training. This further simplifies reward design and avoids reward hacking behavior where policies could learn to ignore the collision penalties.

\begin{figure*}
    \centering
    \includegraphics[width=1\linewidth]{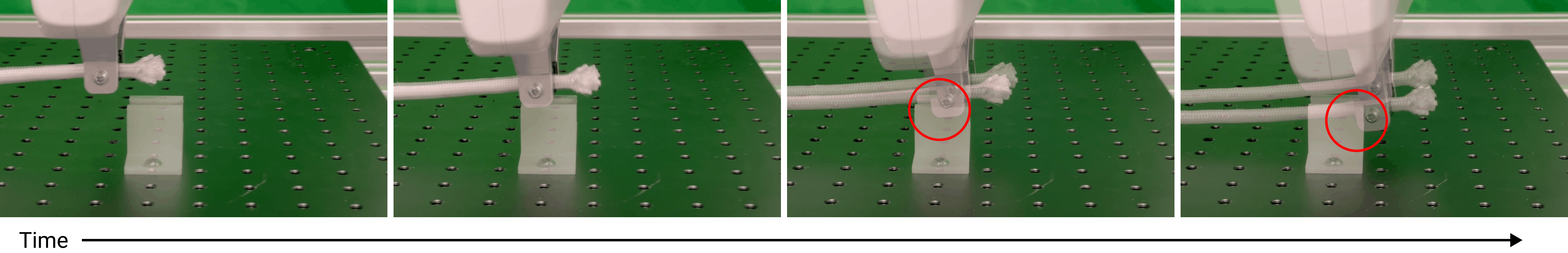}
    \caption{Selected frames of two overlaid videos showing how using SILO with a digital twin that models the harness enables collision avoidance. In one video an RL policy is using SILO with the harness modeled in the simulator. In another video the same policy is using SILO but without the harness modeled in the simulator. The condition without the harness modeled in simulation leads to the robot colliding with the harness and penetrating it. Red circles highlight where the collision would have occurred.}
    \label{fig:silo_collision_avoidance}
\end{figure*}

Figure \ref{fig:silo_collision_avoidance} illustrates this effect by comparing deployment with and without the harness modeled in simulation. When the harness is omitted, the RL policy collides with the physical fixture; when included, SILO prevents the same policy from colliding by construction.

\subsection{Reduced Engineering Overhead}
\begin{figure}
    \includegraphics[width=\linewidth]{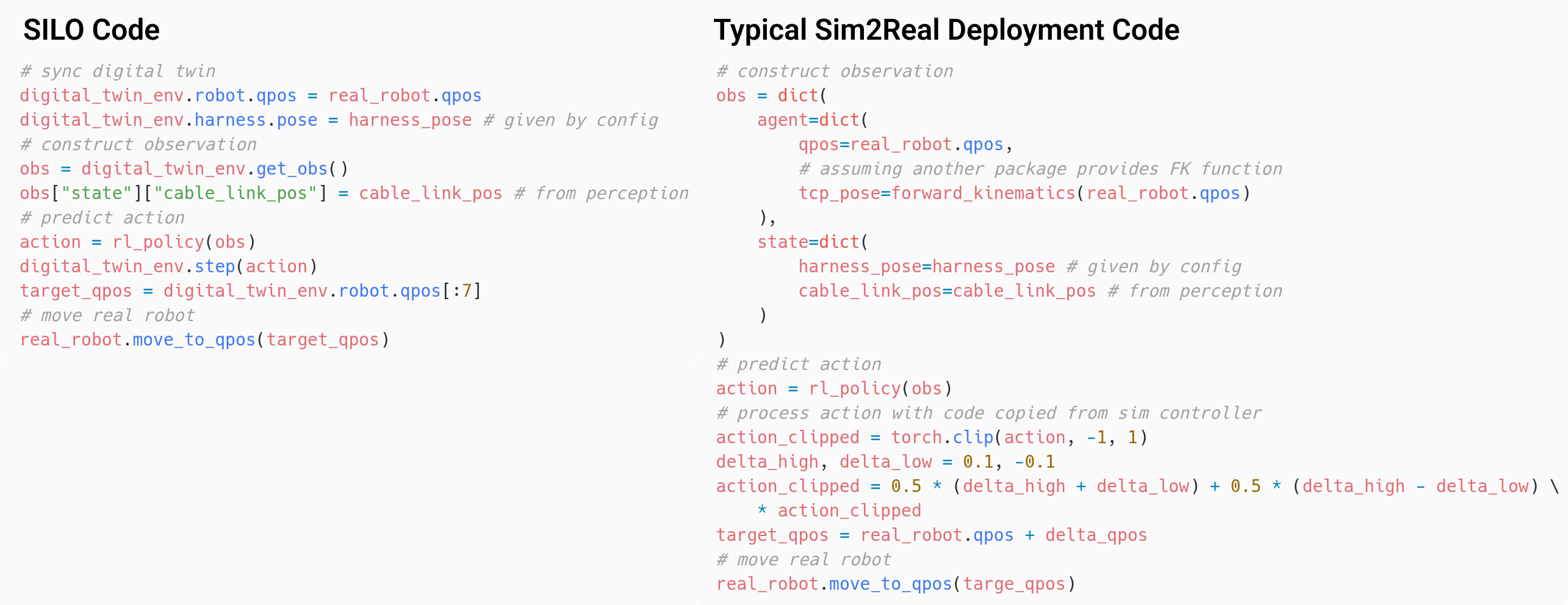}
    \caption{Comparison of code complexity between SILO and a typical sim-to-real deployment setup. This comparison excludes the additional code overhead and time required for controller tuning between simulation and real that typical deployment setups need.}
    \label{fig:code_comparison}
\end{figure}

Typically, sim-to-real workflows require maintaining two separate stacks: one for managing training and one for real world deployment. 
This often leads to duplicated code, increased maintenance burden, longer development cycles, and more implementation bugs.
A comparison of the deployment code alone is shown in Figure \ref{fig:code_comparison}, SILO requires less code in deployment in addition to removing the need for a separate sim-to-real controller tuning algorithm as described in Algorithm \ref{alg:simcontrollertuning}.

SILO addresses these issues by re-using the simulator during deployment, allowing re-use of functionalities such as observation generation or kinematics. Overall, SILO provides several key advantages detailed below:

\textbf{Consistency}: By reusing simulation code during deployment, SILO keeps key functions consistent across training and deployment. For example, when using the simulator's forward kinematics to generate TCP (Tool Center Point) pose data, SILO guarantees the TCP pose generation is consistent.
    
\textbf{Reduced Dependencies}: By reusing the simulator's internal functions (e.g., forward kinematics), SILO avoids the need for additional deployment-time packages for shared functionality. \newaddition{Furthermore, common sim-to-real utilities for controller tuning are no longer necessary, simplifying both the codebase and deployment workflows.}

\textbf{Accelerated Development}: Reducing "duplicate" code minimizes bugs and speeds up the transition from training to testing.

Furthermore, if the real-world camera is calibrated, a matching virtual camera can be instantiated in the simulation. This enables visualizations such as those in Figure \ref{fig:sim-to-real_obs}, providing a direct view of the exact observations the policy processes during deployment. Such transparency is vital for identifying errors in state-estimation predictions that might otherwise impact sim-to-real transfer.

\subsection{Cycle Time Breakdown}

Table \ref{table:cycle_times} reports a breakdown of the cycle time from the main results in Table \ref{table:main_results}.

\textbf{RL Policy} denotes neural network inference time. 

\textbf{Controller} denotes time spent executing real robot motion under primitives or RL actions. 

\textbf{Sim-in-the-Loop} denotes time spent synchronizing the simulator with the real robot and advancing the physics state.

\textbf{Depth Estimation} denotes time to compute a depth image using Foundation Stereo.

\textbf{Cable Segmentation} denotes time to compute a cable mask from an RGB image using SAM2.

\textbf{Observation Processing} denotes time to construct the policy observation, including point cloud segmentation, polyline extraction, and cable point prediction.

\begin{wraptable}{r}{0.42\textwidth}
    \vspace{-1em}
    \small
    \centering
    \begin{tabular}{lcc}
        \toprule
        Component & Time (s) \\
        \midrule
        RL Policy & 0.288$\pm$0.001 \\
        Controller & 41.585$\pm$2.890 \\
        Sim In the Loop & 12.074$\pm$0.149 \\
        Depth Estimation & 12.761$\pm$0.003 \\
        Cable Segmentation & 17.258$\pm$0.282 \\
        Observation Processing & 3.514$\pm$0.034 \\
        \midrule
        Total & 87.480$\pm$2.908 \\
        \bottomrule
    \end{tabular}
    \caption{
    Breakdown of the end-to-end cycle time (in seconds) for routing a cable through three harnesses. Reported values are the mean and standard deviation over 24 real trials. }
    \label{table:cycle_times}
\end{wraptable}
\label{appendix:cycle_times}

Robot execution time dominates overall latency (47.5\%) and varies with harness spacing and initial cable placement; with varying distances required to reach the initial grasp point and traverse between harness poses. Perception modules (depth estimation and cable segmentation) account for a secondary but non-negligible fraction of the cycle time. Given that the perception stack was run on a relatively older RTX 3090, the perception components can be sped up by running newer models like Fast Foundation Stereo \cite{wen2025fast} and SAM3 \cite{carion2025sam} on newer GPUs e.g. RTX 5090. The sim in the loop component accounts for approximately 14\% of the total time. We utilize the GPU-accelerated simulation backend during deployment to ensure consistency with the training environment.

\section{Controller}
\label{appendix:controller} 
\subsection{Real world Controller}
\label{appendix:real-controller}
\newaddition{
For all experiments we control the real robot via the Franky \citep{Schneider_franky_High-Level_Control} library which provides a joint impedance controller that can be controlled by specifying target joint positions. Internally the Franky library relies on the Ruckig package \citep{berscheid2021jerk} to generate motions given a robots joint stiffness and damping parameters. At a low-level, the control frequency between Franky and the real robot is 1kHz. The control frequency of deployed RL policies is set to 10Hz to align with the prediction frequency of the perception pipeline.
}

\subsection{Simulation Controllers}
\label{appendix:sim-controller}
In this paper we primarily use a delta joint position controller. We have also done a few tests with a delta end-effector pose controller, but ultimately found the joint position controller to be more flexible and straightforward for the main results and ablation studies on deployment methods.

The simulation delta joint position controller works as follows to convert policy actions into joint targets

\begin{algorithm}
\caption{Delta Joint Position Controller (Simulation)}
\label{alg:delta-joint-pos-controller}
\begin{algorithmic}[1]
\Require Actions $a$, Current robot joint position $q$
\State Clip $a$ to range $[-1, 1]$ to get $a_\text{clip}$
\State Scale $a_\text{clip}$ to range $[-0.1, 0.1]$ to get $a_\text{scaled}$
\State Define joint targets to be $\hat{q} = q + a_{scaled}$
\State Set simulation joint position targets to $\hat{q}$ and run physics step.
\end{algorithmic}
\end{algorithm}

The simulation delta end-effector pose controller accepts from an agent/policy an action with two components, a desired 3D position and a desired 3D rotation.
\begin{algorithm}
\caption{Delta End-Effector Pose Controller (Simulation)}
\label{alg:delta-ee-pose-controller}
\begin{algorithmic}[1]
\Require Translation action $a_p$, Rotation action $a_R$, Robot joint position $q$, Robot end-effector position $p$, Robot end-effector rotation $R$.
\State Clip $a_p, a_R$ both to range $[-1, 1]$ to get $a_{p,\text{clip}}, a_{R,\text{clip}}$
\State Scale $a_{p,\text{clip}}, a_{R,\text{clip}}$ to range $[-0.1, 0.1]$ to get $a_{p,\text{scaled}}, a_{R,\text{scaled}}$
\State Define $\hat{p} = p + a_{p,\text{scaled}}$
\State Define $\hat{R} = R \cdot e^{[a_{R,\text{scaled}}]_\times}$
\State Solve via Inverse Kinematics (IK) for joint position $\hat{q} = \text{IK}(\hat{p}, \hat{R})$ conditioned on an initial solution $q$.
\State Set simulation joint position targets to $\hat{q}$ and run physics step.
\end{algorithmic}
\end{algorithm}

\subsection{Simulation Controller Tuning}
\label{appendix:sim-controller-tuning}
\begin{figure}[h]
    \centering
    \includegraphics[width=0.9\linewidth]{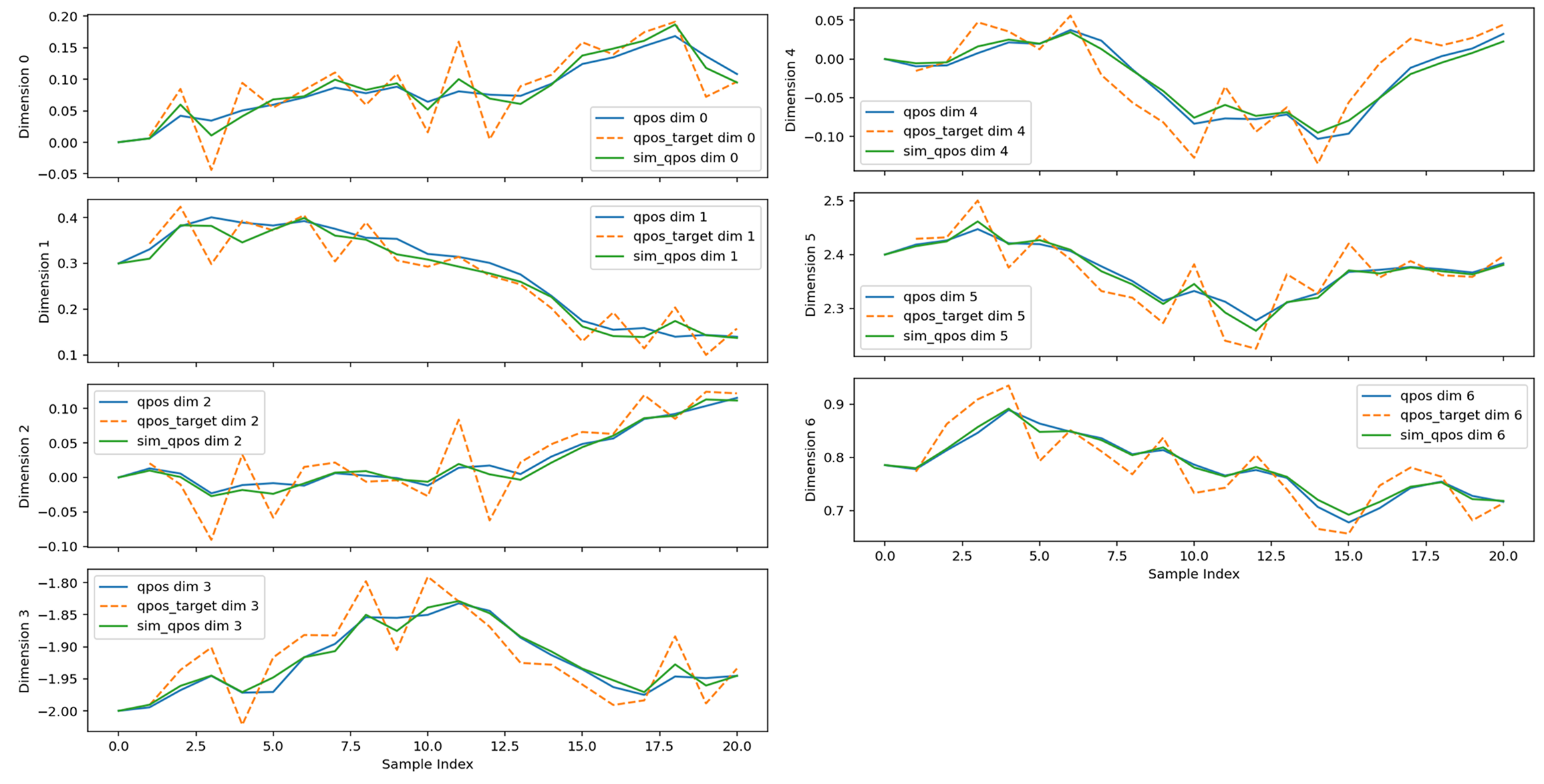}
    \caption{Motion profile of each of the 7 arm joints after calibrating the simulation controller to align with the real robot controller. Dashed orange line indicates the sampled joint target commands. Blue line indicates the ground truth joint positions the real robot achieved. Green line indicates the simulated joint positions.}
    \label{fig:motion_profile}
\end{figure}

\newaddition{
For the baseline approach that does not use SILO and instead uses a conventional sim-to-real deployment method involving manual controller tuning, we determine simulation robot joint stiffness and damping values to match the real world motion profile. The approach involves random sampling simulation parameters, running the same joint targets in simulation that were previously run in the real world, and measuring the tracking error between simulation and real. The method is largely the same as the real-to-sim autotune module used by the Recipe project \citep{lin2025sim} for humanoids+dexterous hands; we present an adapted version in Algorithm~\ref{alg:simcontrollertuning} for clarity. After tuning the simulation controller PD gains, we visualized the motion profile of the simulated robot and real robot in Fig. \ref{fig:motion_profile}. In our setup, we find that on average there is about 0.5 degrees of tracking error, with a worst case of about 1 degree in some steps.
}

\begin{algorithm}
\caption{Aligning Simulation Controller with Real Controller}
\label{alg:simcontrollertuning}
\begin{algorithmic}[1]
\Require Total samples $n$, Total joint targets $m$, Stiffness ranges per robot joint $p_\text{low}, p_\text{high}$, Damping ranges per joint $d_\text{low}, d_\text{high}$, Parallel sim environments $S_i$, Control frequency $c$
\For{$i=1$ to $n$}
    \State $p_i \sim \text{Uniform}(p_\text{low}, p_\text{high})$
    \State $d_i \sim \text{Uniform}(d_\text{low}, d_\text{high})$
    \State Set stiffness and damping parameters $p_i, d_i$ into sim environment $S_i$.
\EndFor
\State $q_\text{targets} \gets$ Sample $m$ joint targets 
\State $q_\text{real} \gets$ Record the achieved joint positions of the real robot when following $q_\text{targets}$ at control frequency $c$.
\State $\text{min\_error} = \infty$
\For{$i=1$ to $n$}
    \State $q_\text{sim} \gets $ Record the achieved joint positions of the simulated robot when following $q_\text{targets}$ at control frequency $c$
    \State $ \text{tracking\_error} \gets $ $\text{MSE}(q_\text{sim}, q_\text{real})$
    \If {$ \text{tracking\_error} < \text{min\_error}$}
        \State $\text{min\_error} \gets \text{tracking\_error}$
        \State $\hat{p} \gets p_i$
        \State $\hat{d} \gets d_i$
    \EndIf
\EndFor
\State \Return $\hat{p}, \hat{d}$
\end{algorithmic}
\end{algorithm}

\section{Sim Design}
\label{appendix:sim-design}
We provide more in-depth details about cable routing simulation environment and formalize the usage of the rigid body APIs. The simulation hyperparameters are shown in Table \ref{table:sim-parameters}.

\subsection{Cable Approximation via Rigid Body Simulation}
For a cable with decimation $d$ and thus $d$ links, $3(d-1)$ joints are modelled, three revolute joints between each adjacent link. Let $q_t \in \mathbb{R}^{3(d-1)}$ be the joint angles of the simulated cable at physics timestep $t$. We define $\hat{q}_t \in \mathbb{R}^{3(d-1)}$ as the joint drive targets of the cable at physics timestep $t$. In the PhysX physics engine, these targets represent the state that the joints are optimized to reach during each step, governed by stiffness and damping parameters. Note that there is a difference between a control step and a physics step. An agent (RL policy in our work) operates at the control step level, but simulation operates at the physics step level. In our case, we use a physics dt of $1/240$ and a control dt of $1/60$, which means we run 4 physics steps per control step.

At the start of an episode, we initialize $\hat{q}_0 = q_0$. For any physics timestep $t > 0$, the targets are updated as: $\hat{q}_t = \hat{q}_{t-1} + \beta (q_t - \hat{q}_{t-1})$ where $\beta$ is a ``plasticity'' coefficient. This joint drive target update essentially moves the current targets closer to the current joint angles of the cable at a magnitude dictated by $\beta$. While our work does not investigate every hyperparameter of our simulation environment due to time/space limitations, we invite future research to look into how to tune these parameters for either a specific cable behavior or tackle even larger-scale generalization.
\subsection{Environment Details}
The simulation environment is comprised of a single robot, a single harness to route through, the cable, and a floor plane representing the workspace table. The robot base is assumed to be at the identity pose.

In our implementation, the floor plane is fixed to a predefined height. The predefined height is tuned by measuring the height of the real world table relative to the robot base. One could improve this by simply randomizing the table height in simulation and training a more robust RL policy.

\begin{table}
\small
\centering
\begin{tabular}{ll}
\hline
\rowcolor{gray!30}
\textbf{Hyperparameter} & \textbf{Value} \\
\hline
\rowcolor{gray!10}
\textbf{Cable} & \\
\hline
Cable Length & 0.375m \\
Cable Decimation & 18 \\
Cable Radius & 0.003 m\\
Cable Joint Drive Stiffness & 1e-2 \\
Cable Joint Drive Damping & 1e-3 \\
Cable Plasticity $\beta$ & 0.7 \\
Cable Y-axis Randomization Range & [-0.2, 0.2]\\
Cable Observed Points & 4 \\
\hline
\rowcolor{gray!10}
\textbf{Harness} & \\
\hline
Harness position range & $[0.3, 0.75] \times [-0.4, 0.4]$ \\
Harness Z-axis rotation range & $[-90^\circ, 90^\circ]$\\
\hline
\rowcolor{gray!10}
\textbf{General Simulation} & \\
\hline
Physics delta time ($dt_p$) & 1/240 \\
Control delta time ($dt_c$)& 1/60 \\
Solver position iterations & 16 \\
Solver velocity iterations & 1 \\
\hline
\end{tabular}
\caption{Hyperparameters for the cable GPU simulation}
\label{table:sim-parameters}
\end{table}

\section{Reinforcement Learning Training}
\label{appendix:reinforcement_learning}

\subsection{Hyperparameters}
The hyperparameters for PPO during any RL training in this work are outlined in Table \ref{tab:hyperparameters}. These are copied verbatim from ManiSkill3's PPO baseline and left largely unchanged apart from the discount factor.

\begin{table}[h]
\small
\centering
    \begin{tabular}{ll}
    \hline
    \rowcolor{gray!30}
    \textbf{Hyperparameter} & \textbf{Value} \\
    \hline
    \rowcolor{gray!10}
    \textbf{PPO} & \\
    \hline
    Discount factor $(\gamma)$ & 0.95\\
    GAE-Lambda $(\lambda)$ & 0.95 \\
    Mini-batch size & 512 \\
    Number of mini batches & 32 \\
    Update epochs & 8 \\
    Advantage normalization & True \\
    Clip coefficient & 0.2 \\
    Clip value loss & False \\
    Entropy coefficient & 0.0 \\
    Value loss coefficient & 0.5 \\
    Max gradient norm & 0.5 \\
    Target KL & 0.1 \\
    \hline
    \rowcolor{gray!10}
    \textbf{Networks and Optimization} & \\
    \hline
    Network Shape of Actor and Critic (MLP) & [256, 256, 256] \\
    Activation & TanH \\
    Learning Rate & 3e-4 \\
    Optimizer & Adam \\
    \hline
    \rowcolor{gray!10}
    \textbf{Environment and Data} & \\
    \hline
    Number of Parallel Environments & 1024 \\
    Episode Horizon & 32 \\
    Steps per environment & 16 \\
    Observation Type & State \\
    Number of cached intermediate reset states & 32768 \\
    Cable Point Observation Noise Scale & 1e-3 \\
    \hline
    \end{tabular}
    \caption{Hyperparameters for all aspects of RL training.}
\label{tab:hyperparameters}
\end{table}
\subsubsection{Training}
\begin{figure}
    \centering
    \includegraphics[width=0.7\linewidth]{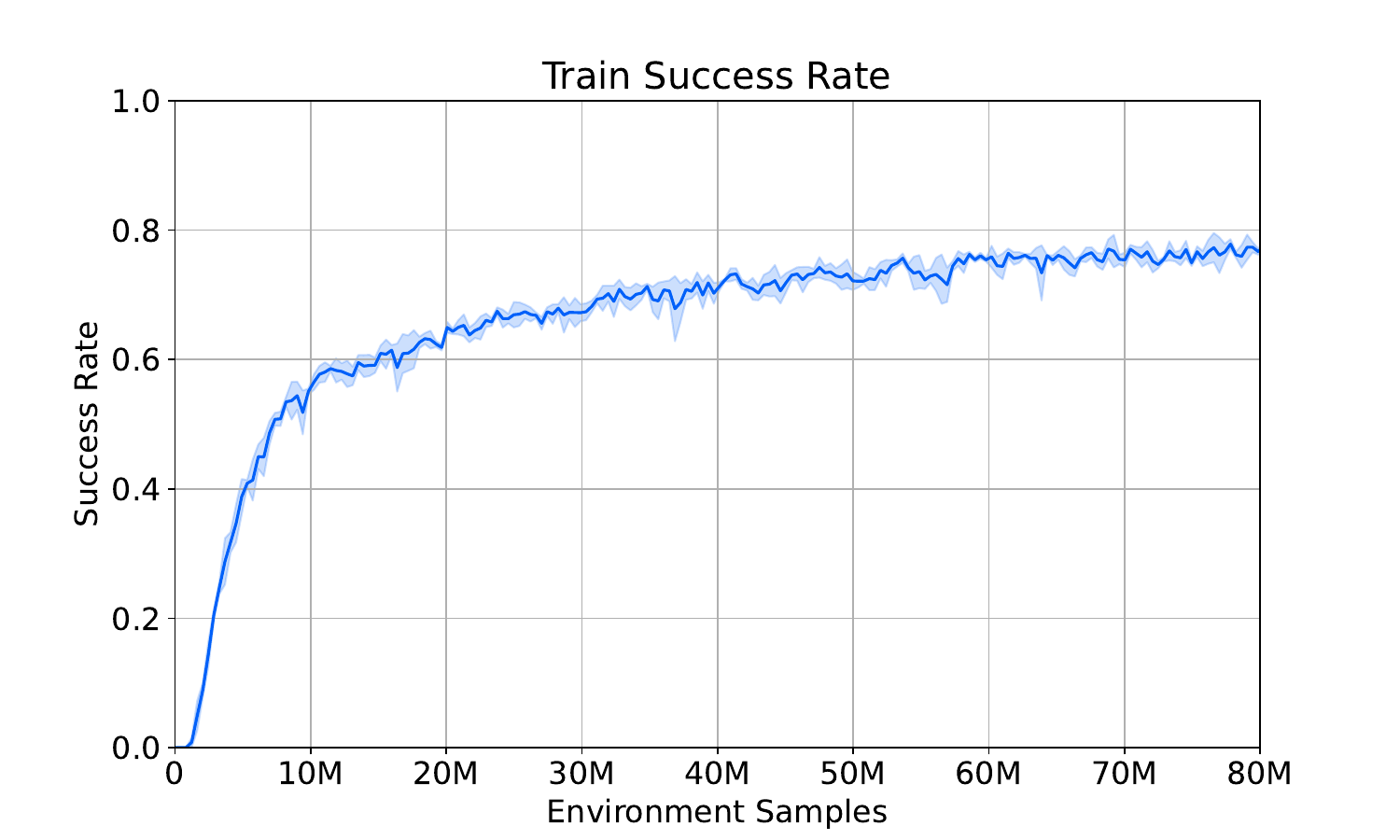}
    \caption{Training success rate over environment samples. Curve is the average of 5 trials, and shaded area is the standard deviation.}
    \label{fig:training_curves}
\end{figure}

Figure \ref{fig:training_curves} shows the RL training curves. Training takes about 5 hours to reach 80 million samples using 1024 parallel environments with 16 steps taken per environment. The training runs were conducted on a server with an NVIDIA RTX 5090 GPU and an AMD Ryzen 9 9950X 16-Core Processor. Note that due to how intermediate states are sampled, it is difficult/near impossible to get close to a 100\% success rate in the environment. We sample harness poses in a very large workspace, some of which are barely reachable by the robot and thus are states that aren't solvable from. One could address this problem by pruning the sampled intermediate states based on a more strict criterion, but we leave that for future work.

\newpage
\section{Real World Setup}
\label{appendix:real_world_setup} 
\begin{figure}
    \centering
    \includegraphics[width=0.8\linewidth]{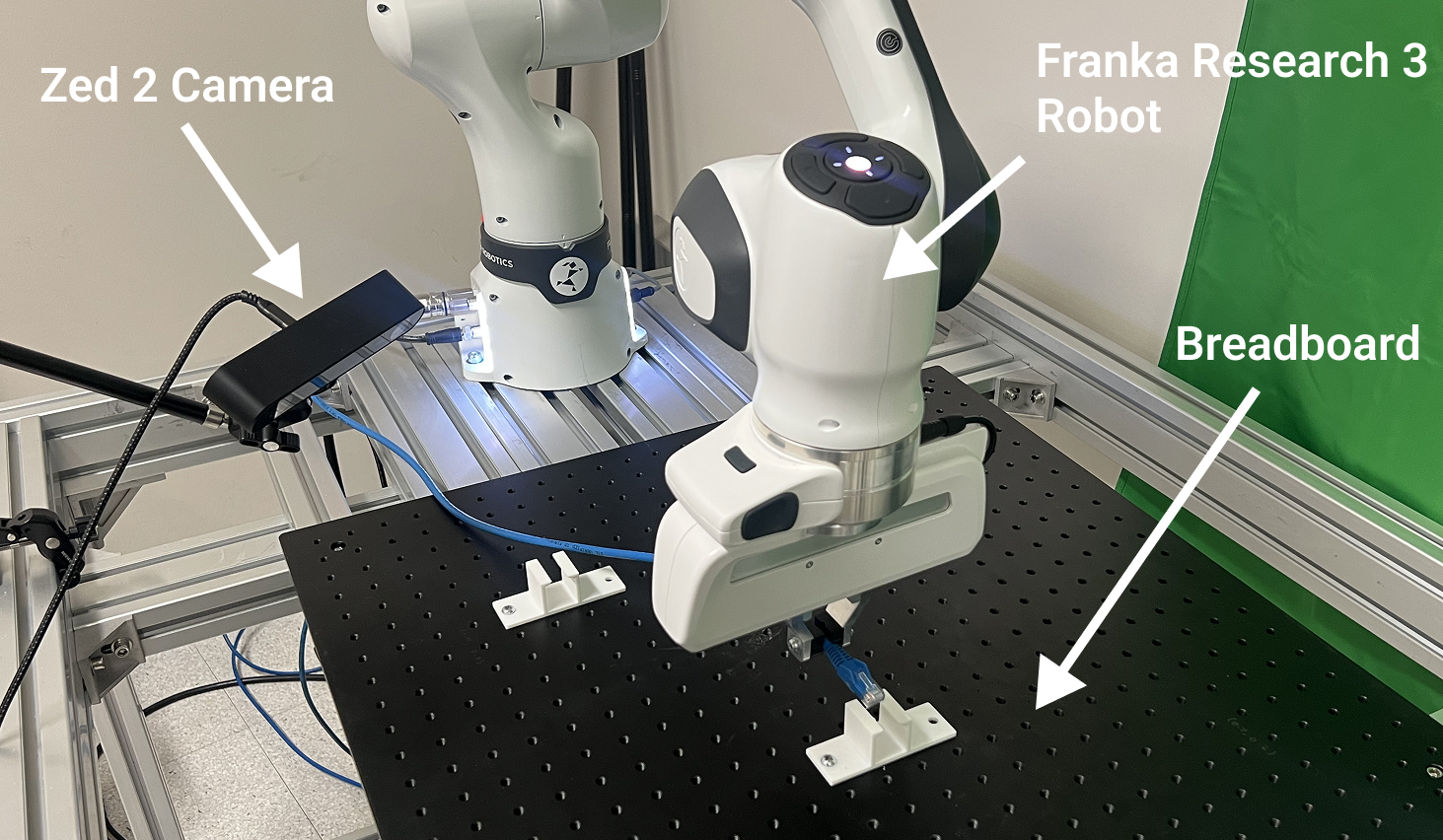}
    \caption{Image of the real world setup for experiments.}
    \label{fig:real_world_setup}
\end{figure}
The real world setup consists of a single robot arm (Franka Research 3 in our experiments), a breadboard fixture used to mount the harnesses in front of the robot, and a third-view stereo RGB camera (Zed 2). Figure \ref{fig:real_world_setup} illustrates the complete setup. The third-view camera is positioned such that it can see all the harnesses and the cable throughout the routing task, enabling reliable state estimation during execution.

\textbf{Cable Properties}
\label{appendix:cable_properties}

\begin{wraptable}{r}{6.5cm}
    \small
    \vspace{-4em}
    \centering
    \begin{tabular}{ccc}
        \hline
         \rowcolor{gray!30} Cable & Thickness (mm) & Material \\
         \hline
         Nylon Rope & 7.9 & Nylon \\
         Ethernet Cable & 6.0 & Plastic\\
         Charger Cable & 6.2 & Plastic \\
         HDMI Cable & 8.0 & Plastic \\
         \hline
    \end{tabular}
    \caption{Properties of the 4 cable types tested. Thickness is measured by using a caliper to measure the diameter of the object.}
    \label{table:cable_properties}
\end{wraptable}

\begin{figure}
    \centering
    \includegraphics[width=0.6\linewidth]{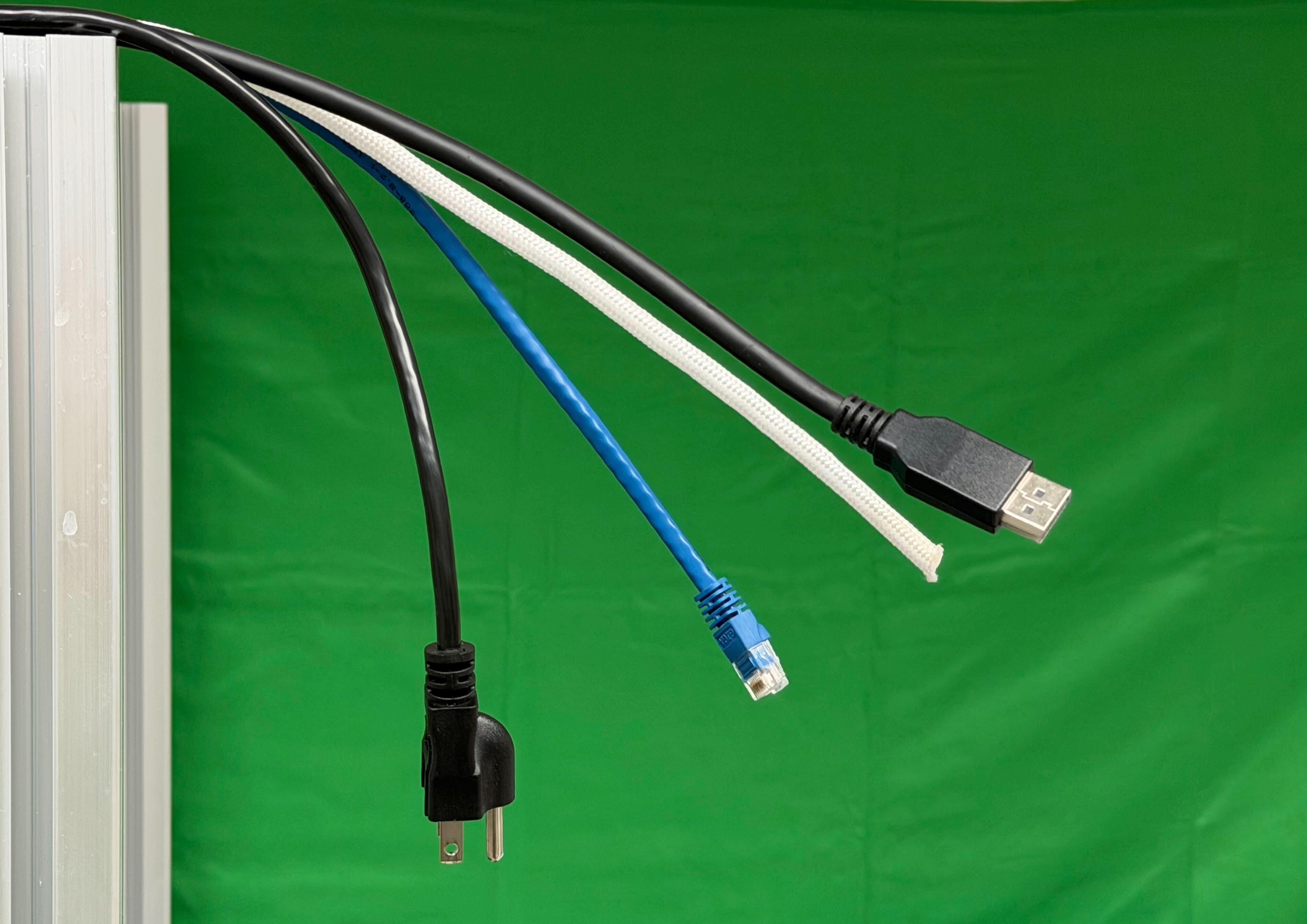}
    \caption{Cantilever test comparing the 4 main cables tested in this paper.}
    \label{fig:cable_cantilever}
\end{figure}

\newaddition{
Table \ref{table:cable_properties} details a number of measured properties of the cables tested in this paper, ranging from a thickness of 6.0mm to 8.0mm, and with different plastic/nylon materials. Figure \ref{fig:cable_cantilever} shows a visual comparison of different cable stiffness profiles via a cantilever test.
}

\section{Additional Results}
\subsection{Learned Behaviors Visualizations}
\label{appendix:learned_behaviors_visualizations}
\begin{figure}
    \centering
    \includegraphics[width=0.95\linewidth]{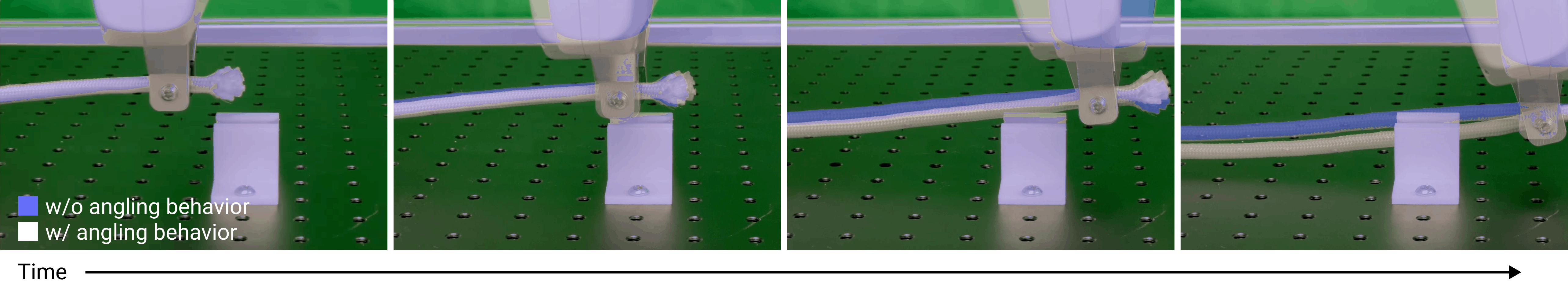}
    \caption{Selected frames of the angling behavior demonstrated by the RL policy when routing through a harness. Frames show an overlay of two videos. The normal-color (white rope and gripper) overlay shows the RL policy changing the gripper orientation as it routes the cable. The blue-tinted overlay shows a scripted baseline that follows the same translational trajectory but maintains a fixed gripper orientation.}
    \label{fig:angling_behavior_time}
\end{figure}
\begin{figure}
    \centering
    \includegraphics[width=1\linewidth]{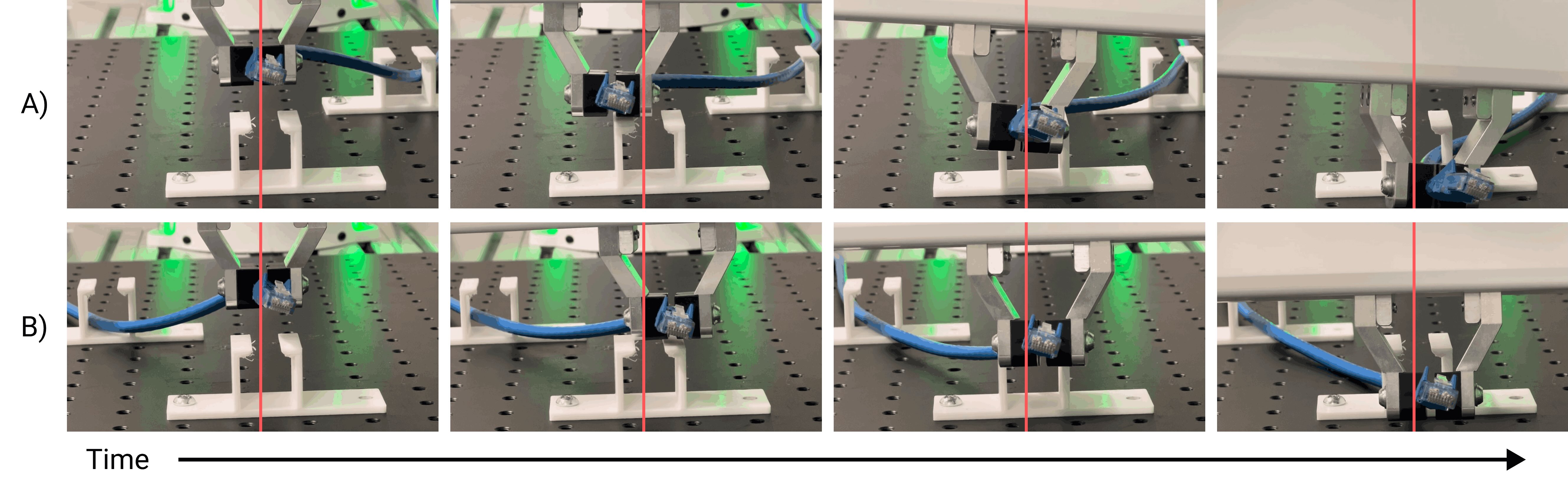}
    \caption{Selected frames of the swinging behavior demonstrated by the RL policy when routing through a harness. Top A) shows the cable curved to the right of the harness being routed through, and the gripper executes a leftward swing. Bottom B) shows the cable curved to the left of the harness being routed through, and the gripper executes a rightward swing.
    A fixed red vertical line is overlaid in each frame as a reference, highlighting the policy’s deliberate lateral displacement of the gripper relative to the harness across time.}
    \label{fig:swinging_behavior_time}
\end{figure}

We find our RL policies learn emergent behaviors in simulation that then transfer to the real world to help facilitate cable routing success. Here, \emph{emergent} means that these behaviors were not explicitly encoded in the reward function or motion primitives. We show step by step frames of two of the behaviors learned by the RL policy in the real world. Videos can also be found on our website that best showcase the subtle behaviors.

\textbf{Angling Behavior:} Fig. \ref{fig:angling_behavior_time} demonstrates the RL policy's \emph{angling} behavior where it adjusts the gripper orientation to face upward which helps push down the cable being routed. This downward motion helps seat the cable into the harness more quickly and reliably. Figure \ref{fig:angling_behavior_time} overlays two trajectories: the RL policy in natural color and a comparative baseline in blue tint. The baseline (in blue) utilizes a scripted policy that replicates the RL policy's translational path but maintains a fixed gripper orientation. Notably, the RL policy depresses the cable approximately 8~mm further than the baseline—equivalent to the cable's diameter. At the end of the control horizon, the baseline policy without the angling behavior fails to fully seat the cable, whereas the RL policy achieves successful routing. 

\textbf{Swinging Behavior:} Fig. \ref{fig:swinging_behavior_time} shows selected frames of the emergent swinging behavior learned by the RL policy. The robot executes lateral movements in response to the curvature of the cable relative to the harness. When the cable is curved to the right of the harness (top row, A), the policy swings the gripper leftward around the fixture; when the cable is curved to the left (bottom row, B), the policy swings to the right. Notably, this behavior is not explicitly encoded in the reward function and emerges from interaction with the simulated dynamics during training.

\subsection{Observation Space Ablation}
\begin{wraptable}{r}{0.43\textwidth}
    \vspace{-2em}
    \small
    \centering
    \begin{tabular}{cccc}
        \toprule
         Observed Points & 1 & 2 & 4 \\
         \midrule
         Success Rate & 1/20 & 11/20 & \textbf{15/20} \\
         \bottomrule
    \end{tabular}
    \caption{Success rate of policies trained with different numbers of observed cable points.}
    \label{table:observation_space_ablation}
    \vspace{-1.5em}
\end{wraptable}

We were motivated to perform a sanity check of whether the cable deformation truly mattered to the RL policies we trained. To do so, we studied how the observation space affects learned behaviors and sim-to-real transfer. Specifically, we vary the number of cable points $\{1, 2, 4\}$ provided to the policy while keeping all other aspects of training fixed. Table~\ref{table:observation_space_ablation} shows that observing a single cable point is insufficient, leading to nearly deterministic behavior that ignores cable curvature. Observing two points significantly improves performance and enables both angling and swinging behaviors visualized and described in Section \ref{appendix:learned_behaviors_visualizations}, while four points yield the highest success rate. 
We found that exceeding four points introduces diminishing returns and increases sensitivity to visual occlusions; in the dense workspace of the harness sequence, additional points were frequently obstructed, leading to a distribution shift that hindered zero-shot transfer.
These results indicate that capturing cable curvature, rather than position alone, is critical for robust reactive routing. It further underlines how the simulated cable model captures the real world cable behaviors well enough to facilitate sim-to-real transfer.

\subsection{Failure modes}
\label{appendix:example_failures}
\begin{figure}
    \centering
    \includegraphics[width=1\linewidth]{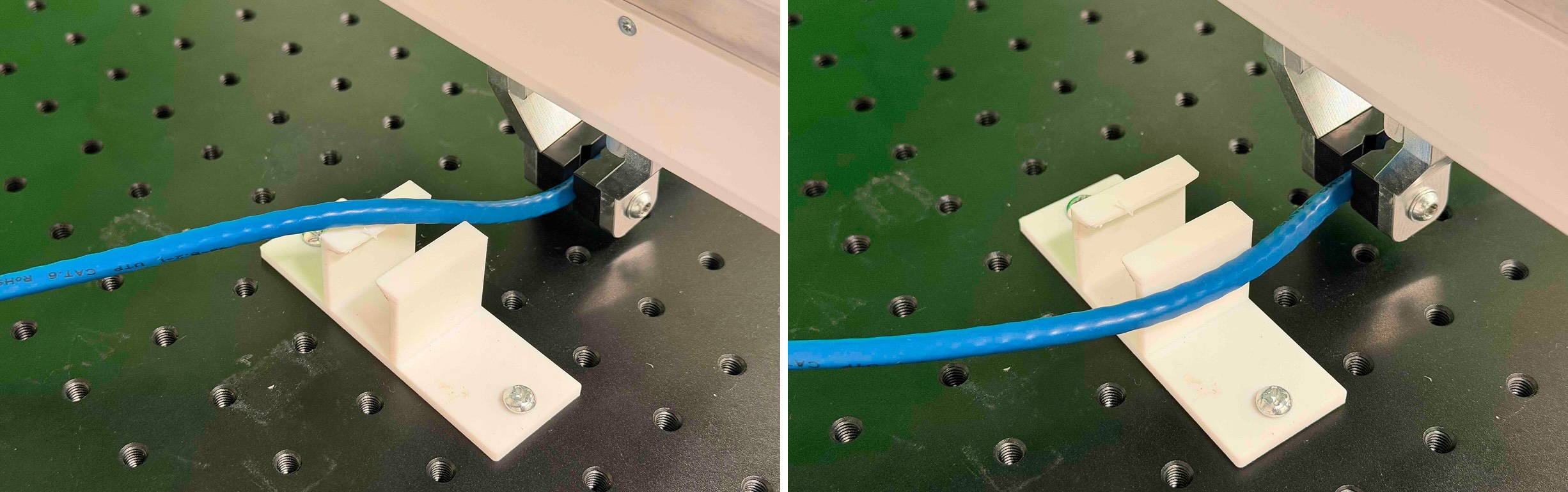}
    \caption{Two common failure types during execution where the RL policy fails to route completely. Left image shows what happens when the cable is snagged onto the harness and fails to route through. Right image shows when the policy completely misses the harness when routing. }
    \label{fig:failure_modes}
\end{figure}

There are a few common failure modes of our system when executing various RL policies (baselines as well as our best setups) when routing through harnesses shown in Fig. \ref{fig:failure_modes}. 

The first is when the cable gets snagged onto a part of the harness and the policy fails to make the correct fine-grained adjustments to route through completely. This failure is the most common failure mode and is the only failure that occurs when using SILO and our best RL policy. It often happens if a policy is not trained with sufficient randomization and/or the cable entry angle is too large, making it difficult to route through.

The second is when the policy completely misses the harness altogether and the cable goes outside of the harness. Some baselines with insufficient randomization will exhibit this failure.

The last failure mode is when the robot enters a ``reflex mode'' that triggers whenever there is hard contact between the robot and an object (the harness in our case). The reflex mode is a safety measure from the Franka robot arm that stops the robot whenever there is too much force due to contacts. This failure mode only occurs when we are not using SILO and do not use contact penalties in our RL training.

\end{document}